\documentclass[11pt]{article}

\usepackage[english]{babel}

\usepackage[letterpaper,top=2cm,bottom=2cm,left=3cm,right=3cm,marginparwidth=1.75cm]{geometry}

\usepackage{amsmath, amssymb, amsfonts, bm}
\usepackage{graphicx}
\usepackage{float}
\usepackage[colorlinks=true, allcolors=blue]{hyperref}
\usepackage{rotating, threeparttable, multirow, multicol, makecell, booktabs}
\usepackage{array}
\usepackage{authblk}
\usepackage{indentfirst}
\usepackage{subcaption}
\usepackage{cite, natbib}
\usepackage{threeparttable}%
\usepackage[title]{appendix}
\usepackage{soul}
\usepackage{setspace}
\title{Uncertainty-Aware Data-Based Method for Fast and Reliable Shape Optimization}

\author[1]{Yunjia Yang \thanks{yyj980401@gmail.com}}
\author[1]{Runze Li}
\author[1]{Yufei Zhang}
\author[1]{Haixin Chen \thanks{Corresponding Author:  chenhaixin@tsinghua.edu.cn}}

\affil[1]{School of Aerospace Engineering, {Tsinghua University, Beijing 100084, China}}

\begin{document}

\maketitle

\abstract{Data-based optimization (DBO) offers a promising approach for efficiently optimizing shape for better aerodynamic performance by leveraging a pretrained surrogate model for offline evaluations during iterations. However, DBO heavily relies on the quality of the training database. Samples outside the training distribution encountered during optimization can lead to significant prediction errors, potentially misleading the optimization process. Therefore, incorporating uncertainty quantification into optimization is critical for detecting outliers and enhancing robustness. This study proposes an uncertainty-aware data-based optimization (UA-DBO) framework to monitor and minimize surrogate model uncertainty during DBO. A probabilistic encoder-decoder surrogate model is developed to predict uncertainties associated with its outputs, and these uncertainties are integrated into a model-confidence-aware objective function to penalize samples with large prediction errors during data-based optimization process. The UA-DBO framework is evaluated on two multipoint optimization problems aimed at improving airfoil drag divergence and buffet performance. Results demonstrate that UA-DBO consistently reduces prediction errors in optimized samples and achieves superior performance gains compared to original DBO. Moreover, compared to multipoint optimization based on full computational simulations, UA-DBO offers comparable optimization effectiveness while significantly accelerating optimization speed.}




\maketitle

\section{Introduction}\label{sec1}

Advancements in computer technology have continually reshaped aerodynamic design methodologies. In recent years, significant efforts have focused on applying machine learning (ML) techniques to accelerate optimization processes \citep{brunton_data-driven_2021,li_machine_2022}. A prevalent strategy involves using ML to construct surrogate models for aerodynamic performance or flow field predictions. Among the notable frameworks is data-based optimization (DBO) \citep{li_data-based_2019}, also referred to in some literature as interactive optimization \citep{du_rapid_2021}. In DBO, a surrogate model is pretrained on a large, diverse dataset that broadly covers the design space of interest and is subsequently deployed offline within an optimization loop, largely or entirely eliminating the need for CFD evaluations during optimization.

Compared to efficient global optimization (EGO) where surrogate models are iteratively refined during optimization, DBO offers both advantages and challenges due to its offline nature. On the one hand, DBO can significantly accelerate optimization since no additional simulations are required during iterations, which becomes particularly beneficial in multipoint optimization tasks\citep{li_data-based_2019, yang_fast_2024}. Plus, pretrained models in DBO are highly reusable, enabling the same model to be applied across multiple downstream optimization problems within a similar domain. This amortizes the high cost of generating the pretraining dataset \citep{renganathan_enhanced_2021, iliadis_dnn-driven_2023, yang_fast_2024}. These strengths have enabled successful applications of DBO to airfoil and wing optimization \citep{yang_fast_2024, li_data-based_2021}, nozzle design \citep{yang_interactive_2023}, and aircraft configuration optimization \citep{secco_artificial_2017}.

However, a key challenge persists: DBO’s reliance solely on offline models during optimization introduces risks to result reliability. Without access to ground-truth evaluations to correct the offline surrogate model, optimization may be misled when encountering samples outside the distribution of the training database, leading to significant errors.

To mitigate this risk, incorporating uncertainty quantification (UQ) offers a promising direction to enhance the robustness of DBO. Nevertheless, existing UQ-aware optimization methods are predominantly developed in the context of EGO, where uncertainty estimates are primarily used to guide exploration and active sampling through acquisition functions\citep{gaudrie_modeling_2020, laurenceau_building_2008, li_surrogate-based_2019}. In contrast, such methodology is unavailable in DBO, since no new data can be acquired during optimization. Consequently, uncertainty must be used differently: not to guide sampling, but to inform the optimization process itself by regulate exploitation and prevent overconfidence in unreliable predictions. This subtle yet crucial distinction has received limited attention in the literature. 

Furthermore, DBO typically involves large and high-dimensional datasets for surrogate model pretraining, which poses scalability challenges for GP-based methods \citep{bouhlel_scalable_2020,liu_when_2020}. Although recent advances in UQ for deep neural networks, such as ensemble methods \citep{yang_scalable_2022}, Monte Carlo dropout \citep{mufti_shock_2024}, Bayesian networks \citep{anhichem_bayesian_2025}, and generative models \citep{jabon_aerodynamic_2024,saetta_uncertainty_2024, liu_uncertainty-aware_2024}, show promise in handling large-scale problems, their application within DBO frameworks remains largely unexplored.

To address these gaps, this work proposes an uncertainty-aware data-based optimization (UA-DBO) framework. The core idea is to explicitly incorporate model confidence into the optimization objective, enabling the optimizer to balance performance improvement against the risk associated with unreliable surrogate predictions. This concept is inspired by the robust optimization, though in our case, the uncertainty originates from the pretrained surrogate model, rather than from variability in system inputs. 

The challenge of efficiently estimating prediction uncertainty from pretrained models is also studied. To this end, a Gaussian stochastic encoder decoder (GS-ED) model is proposed. It is built upon the encoder-decoder (ED) architectures that are commonly used for prediction in DBO, and by incorporating the variational approximation techniques.

The effectiveness of the proposed UA-DBO framework is demonstrated through two airfoil multipoint aerodynamic optimization problems, where DBO has shown particular strength.

\section{Uncertainty-aware data-based optimization framework}\label{sec2}

\subsection{Overall framework}

\begin{figure}[htbp]
    \centering
    \begin{subfigure}{\textwidth}
        \centering
        \includegraphics[width=0.55\linewidth]{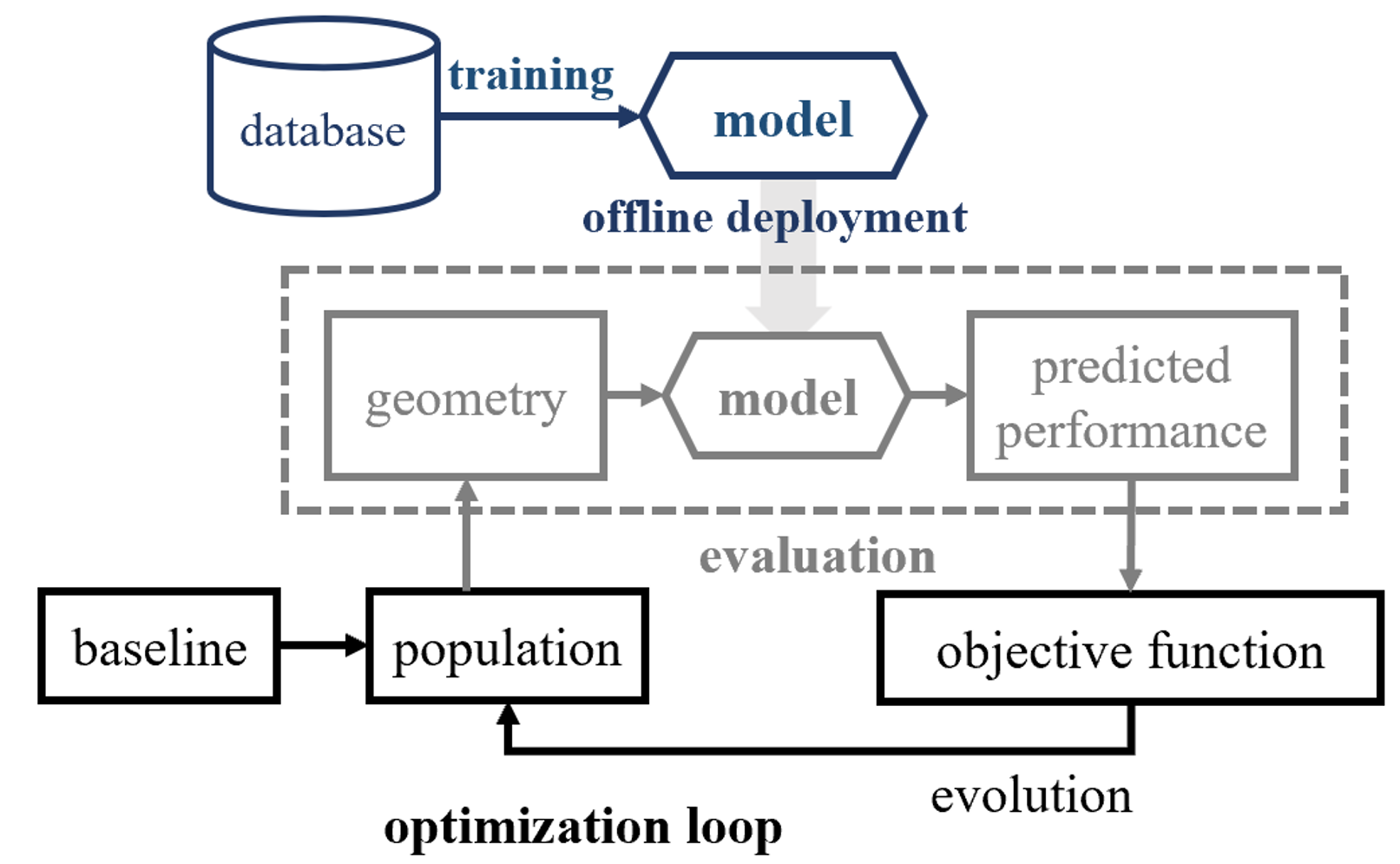}
        \caption{DBO}
    \end{subfigure}

    \begin{subfigure}{\textwidth}
        \centering
        \includegraphics[width=0.55\linewidth]{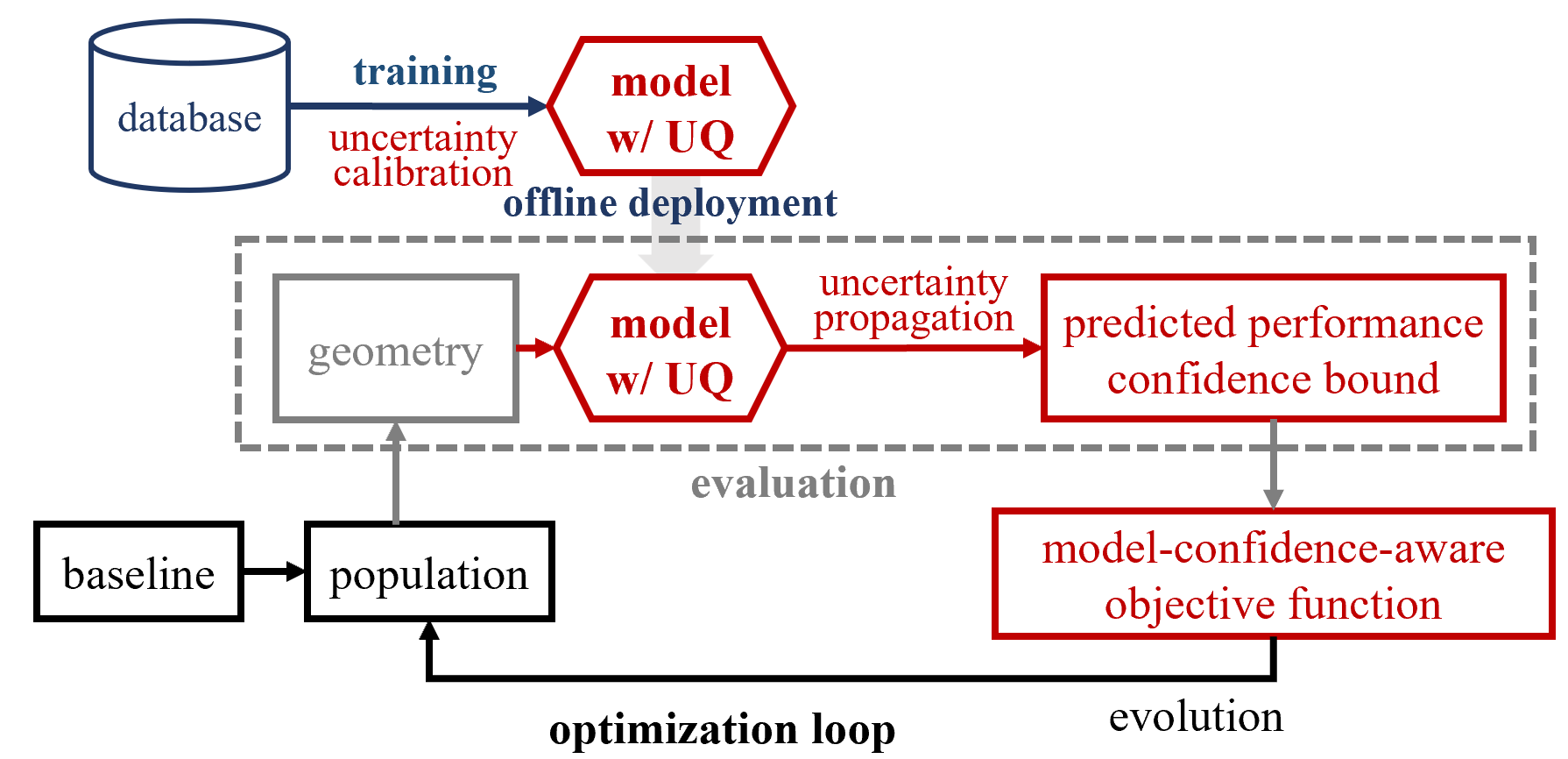}
        \caption{UA-DBO}
    \end{subfigure}
    \caption{Comparison of the DBO and UA-DBO frameworks}
    \label{fig:framework}
\end{figure}

Fig. \ref{fig:framework} shows a comparison of the conventional DBO and the proposed UA-DBO frameworks. In DBO, the model is trained to predict the performance without any extra information on the uncertainty or reliability of the prediction, and the optimization objective function is only based on these predictions. The optimization (in this paper, it is an evolution algorithm) then uses the objective functions in iterations until it obtains the optimal result, which raises the risk that model prediction errors misguide optimization.

UA-DBO leverages UQ to mitigate this risk. The surrogate model is pretrained to predict both the performance metrics and corresponding uncertainty indicators for each prediction, such as the confidence bounds. This necessitates the introduction of uncertainty quantification, propagation, and calibration techniques. During optimization, a model-confidence-aware objective function is employed to jointly maximize performance and minimize uncertainty at optimal locations to achieve robust results. 

The following sections describe the construction of the model-confidence-aware objective function, and the methods to quantify, propagate, and calibrate the model uncertainty.

\subsection{Model-confidence-aware objective function}\label{sec:objintro}

Given an optimization problem to minimize performance metrics $Y(x)$ with respect to the design variable $\bm x$. DBO pretrains a surrogate model $\mathcal F_D$ in advance to learn the deterministic mapping relation between the design variable $\bm x$ and the performance variable $Y(x)$, $\hat Y(x) = \mathcal F_D(x)$. Then the surrogate model is offline deployed in optimization problems to find the optimal value by

\begin{align}
    & \min_{\bm x} \hat Y(x) =\min_{\bm x} \mathcal F_D(\bm x),\\
    &s.t.~~ \mathcal {G} (x) \ge 0 \nonumber
\end{align}

In UA-DBO, a probabilistic model replaces the original deterministic model to give the confidence interval of the model-predicted performance metric. For a minimization problem, the upper bound (UB) of confidence interval, $UB(Y(x))$, is predicted with a surrogate model $\mathcal F_{P, UB}$. When offline applying it in optimization, the model-confidence-aware objective function uses the $UB(Y(x))$ instead of the mean prediction $Y(x)$:

\begin{align}
    &\min_{\bm x} UB(Y(x)) =\min_{\bm x} \mathcal {F}_{P, UB}(\bm x)\\
    &s.t.~~ \mathcal {G} (x) \ge 0 \nonumber
\end{align}

This strategy ensures that the actual optimized performance $Y^{*(\mathrm{CFD})}$ is regulated to be better (lower) than the predicted upper bound, thus reducing the risk of optimization being misled by model errors. 

\begin{figure}[htbp]
    \centering
    \includegraphics[width=1\linewidth]{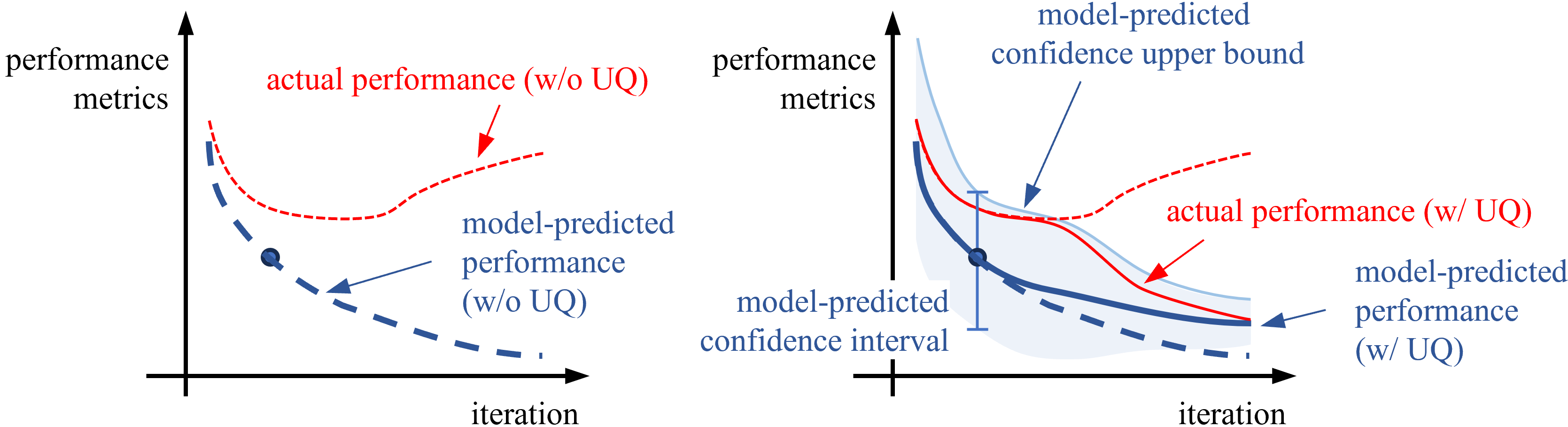}
    \caption*{
        \begin{minipage}{0.4\linewidth}
            \centering
            \small (a) DBO
        \end{minipage}
        \hfill
        \begin{minipage}{0.5\linewidth}
            \centering
            \small (b) UA-DBO
        \end{minipage}
    }
    \caption{Demonstration of possible optimization trajectory with and without UQ}
    \label{fig:path}
\end{figure}

Fig. \ref{fig:path} illustrates the impact of incorporating UQ into DBO. In both subfigures, the blue dashed lines represent the surrogate model’s predicted performance metrics during optimization, while the red dashed lines show the actual performance metrics of the model-predicted best design at each iteration. Without UQ, as shown in Fig. \ref{fig:path}a, the optimization may converge toward a region where the surrogate model predicts good performance, but the actual performance is poor due to model errors. In contrast, Fig. \ref{fig:path}b demonstrates a trajectory incorporating UQ, where, assuming accurate confidence intervals, the actual performance (red solid line) remains within the confidence bounds, preventing error accumulation and misleading optimization.


The core idea of the proposed objective function lies in its applicability to the DBO paradigm, in which a surrogate model is first pre-trained and then applied offline for optimization. As a result, the uncertainty involved in this process can be interpreted from two perspectives. From the surrogate modeling perspective, the uncertainty captured by the model originates from limited training data and model approximation, and is therefore epistemic in nature. However, once the model is deployed for optimization, it is treated as a fixed and external system from the optimizer’s perspective, which behaves functionally like an aleatory uncertainty.

This interpretation naturally allows the optimization problem to be formulated within the framework of robust optimization, where system uncertainties are treated as exogenous disturbances and the objective is to seek solutions that remain effective under worst-case or high-probability realizations of model error. When the surrogate model output is represented by a predictive distribution characterized by a mean $\mu(\bm{x})$ and a standard deviation $\sigma(\bm{x})$, the upper bound of the confidence interval at confidence level $\alpha$ can be directly inferred from these distributional parameters as
\begin{equation}\label{eqn:ub}
    \mathrm{UB}(Y(\bm{x})) = \mu(\bm{x}) + \frac{t_{\alpha, N_s-1}}{\sqrt{N_s}} \, \sigma(\bm{x}),
\end{equation}

where $t_{\alpha, N_s-1}$ denotes the Student’s $t$-distribution with confidence level $\alpha$ and $N_s-1$ degrees of freedom, and $N_s$ is the number of samples used to estimate the predictive statistics.

Under this formulation, the proposed model-confidence-aware objective function is equivalent to a weighted-sum expectation–variance trade-off problem:
\begin{equation}
    \min_{\bm x} \left(\mu(Y) + w \cdot \sigma(Y)\right), \quad Y = \mathcal F(x),
\end{equation}

with the weighting factor given by $w = \frac{t_{\alpha, N_s-1}}{\sqrt{N_s}}$. In previous studies, weighted-sum formulations have been widely adopted due to their simplicity and effectiveness; however, selecting an appropriate weighting factor remains a critical and often problem-dependent challenge\citep{yao_review_2011, du_optimum_2019}. The proposed approach reformulates this weight-selection problem as the choice of a confidence level $\alpha$, which endows the trade-off parameter with a clear probabilistic interpretation and enables more intuitive and principled control over the degree of risk aversion in offline surrogate-based optimization.

\subsection{Uncertainty quantification for the machine learning model}\label{subsec2}

\begin{figure}[htbp]
    \centering
    \begin{subfigure}{\textwidth}
        \centering
        \includegraphics[width=0.45\linewidth]{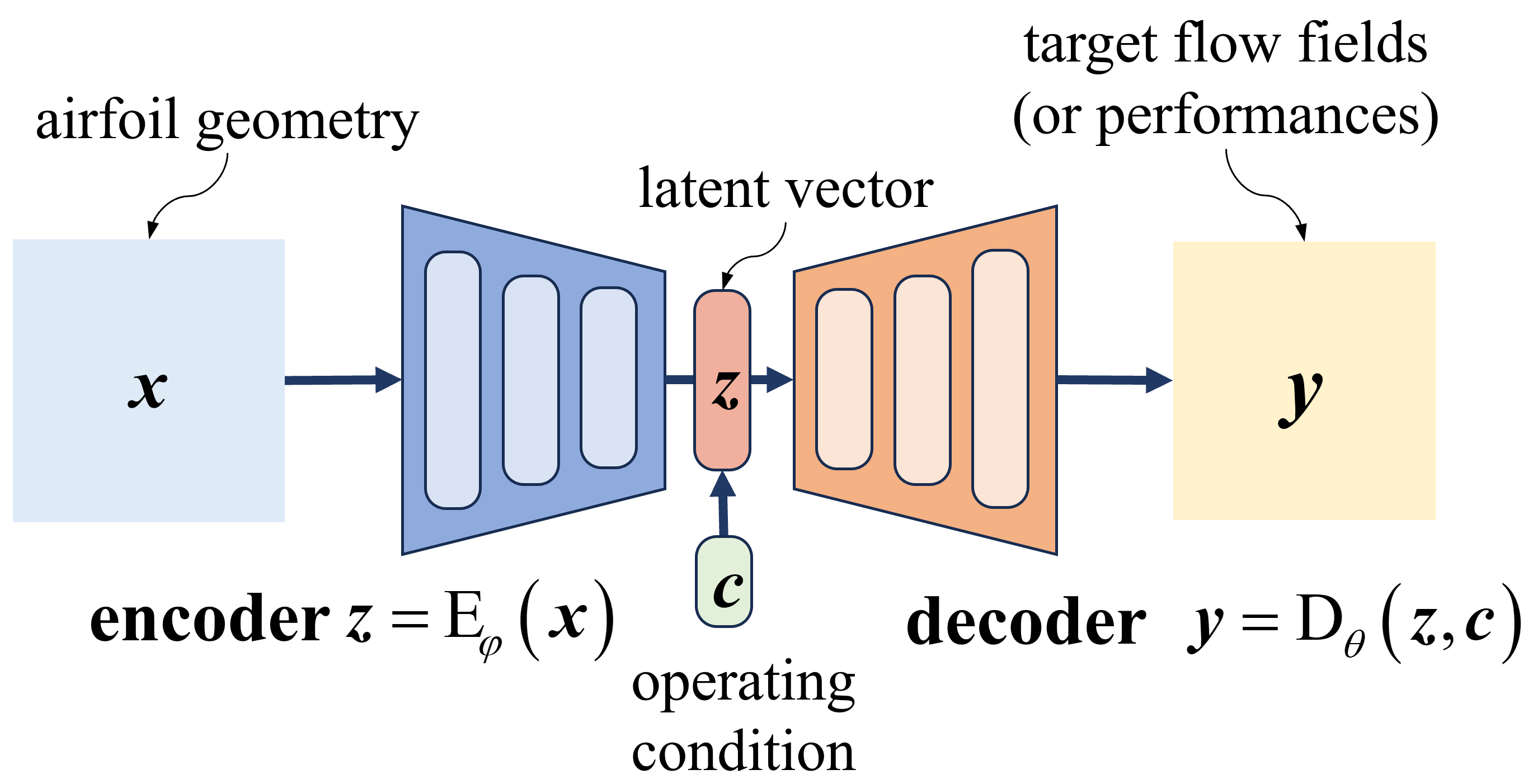}
        \caption{Deterministic encoder-decoder model}
        \label{fig:dedmodel}
    \end{subfigure}

    \begin{subfigure}{\textwidth}
        \centering
        \includegraphics[width=0.47\linewidth]{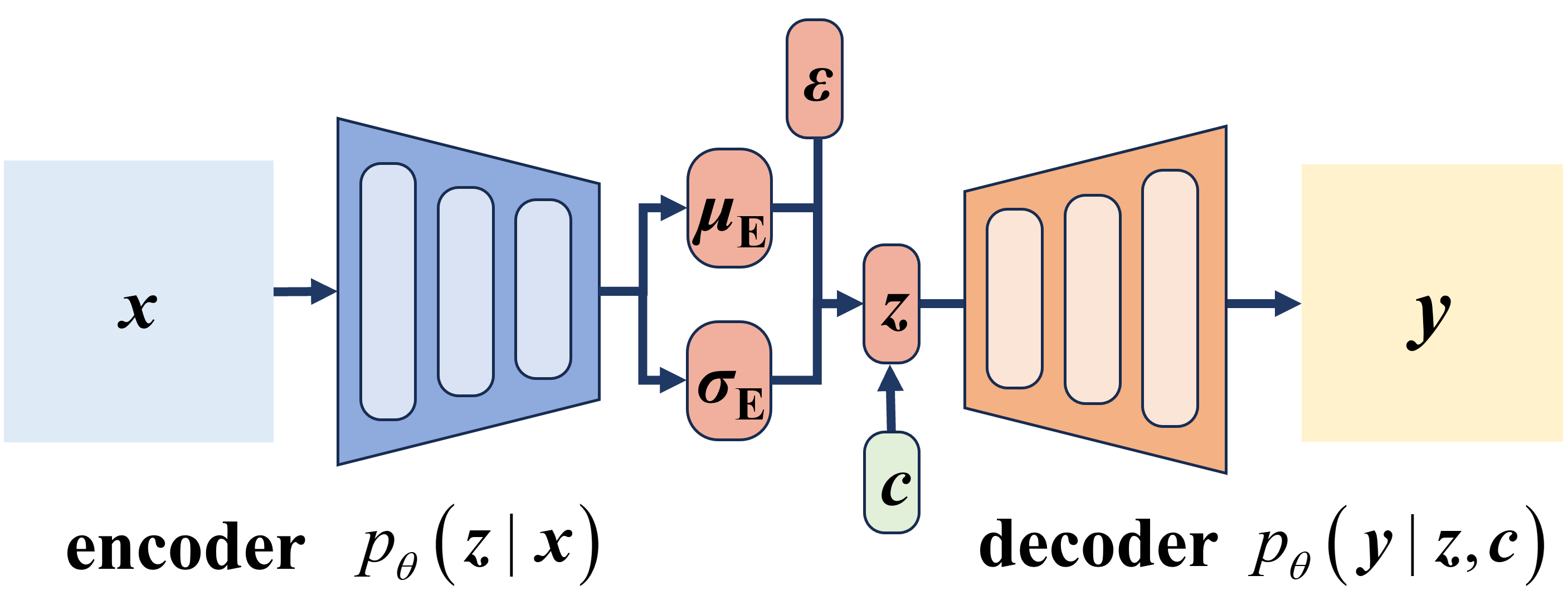}
        \caption{Gaussian stochastic encoder decoder model (GS-ED)}
        \label{fig:pgsm}
    \end{subfigure}
    \label{fig:mlmodels}
    \caption{Architectures of the proposed machine-learning models}
\end{figure}

Fig. \ref{fig:dedmodel} illustrates the deterministic surrogate model commonly adopted in the DBO framework, which is based on an encoder–decoder (ED) architecture. The model takes the geometry $\bm{x}$ and operating conditions $c$ as inputs. The encoder maps these inputs into a low-dimensional latent representation $\bm{z}$, which is concatenated with $c$ and subsequently passed through the decoder to predict either the flow field or the performance metric $\bm{y}$.

In the UA-DBO framework, we extend this architecture by introducing the Gaussian stochastic encoder–decoder (GS-ED) model, designed to provide UQ. The GS-ED model is inspired by the commonly used variational autoencoder (VAE) proposed by \cite{kingma_auto-encoding_2014}. As shown in Fig. \ref{fig:pgsm}, it retains the ED structure while replacing the deterministic latent space with a probabilistic one. UQ is realized by training the model with a variational approximation and estimating uncertainty via Monte Carlo sampling from the probabilistic latent space. This modification is designed to be minimal with respect to the existing DBO pipeline, while it is worth to mention that our proposed UA-DBO framework is compatible other UQ methods.

\subsubsection{Training of GS-ED model}

The key difference from the deterministic ED lies in the latent representation: the encoder now predicts the conditional distribution of the latent $q_\phi(\bm z | \bm x)$, modeled as a Gaussian parameterized by $\left(\bm \mu_E(\bm x^{(i)}), \bm \sigma_E(\bm x^{(i)})\right)$. The decoder network, similarly, now predict the conditional posterior $p_\theta(\bm y|\bm z, c)$.

Analogous to the VAE, the loss function of the GS-ED model is derived by maximizing the evidence likelihood using variational approximation. Yet, unlike the VAE, which is primarily designed for dimensionality reduction, the GS-ED model is tailored to the prediction task required in UA-DBO. Specifically, it maximizes the conditional evidence likelihood $\log p_\theta(\bm y | \bm x, c)$, rather than the plain one $\log p_\theta(\bm y)$. As detailed in Appendix \ref{secA1}, the resulting loss function is

\begin{equation}
    \mathcal L_\mathrm{GS-ED}(\bm y;\theta,\phi)=-\mathbb E_{z\sim q_\phi(\bm z | \bm y)}\log p_\theta(\bm y|\bm z, c)
\end{equation}

During experiments, we observed that relying solely on the likelihood loss could result in large prediction errors in results and especially in uncertainty estimations. A significant reason for this is that the learned latent distributions are no longer guaranteed to be Gaussian, resulting in a degradation in predictive accuracy. To mitigate this, the KL divergence between the conditional prior $q_\phi(\bm z | \bm x)$ and the standard normal distribution $KL\left(q_\phi(\bm z | \bm x), \mathcal{N}(0, \mathbf{I})\right)$ was reintroduced as a regularization term, weighted by a hyperparameter $\beta$.

To enable gradient-based optimization, the reparameterization trick is employed. For each input $(\bm{y}^{(i)};\bm{x}^{(i)},c^{(i)})$, $N_l$ latent variables  $\left\{z^{(i,l)}\right\}_{i=1,\cdots,N_l}$ are sampled from the recognition distribution $q_\phi(\bm z | \bm y)$. Assuming the posterior distribution $p_\theta(\bm y | \bm z)$ follows a Gaussian with mean $\bm \mu_D(\bm z^{(i,l)})$ predicted by the decoder and identity covariance, the loss for each training instance becomes:

    \begin{align}
    \mathcal L_\mathrm{GS-ED}(\bm y^{(i)};\theta,\phi)=\frac{1}{N_l}\sum_{l=1}^{N_l}&\frac{1}{2}\left\Vert\bm y^{(i)}-\bm \mu_D(\bm z^{(i,l)}, c^{(i)})\right\Vert_2^2 \nonumber \\
    & -\frac{1}{2} \big(1 + \log \bm \sigma_E^2(\bm x^{(i)}) - \bm \mu_E^2(\bm x^{(i)}) - \bm \sigma_E^2(\bm x^{(i)})\big)
    \label{eqn:trainloss}
\end{align}
\noindent where $\bm z^{(i,l)}$ is sampled from $\mathcal N \left(\bm \mu_E(\bm x^{(i)}, \bm \sigma_E(\bm x^{(i)})\right)$ denotes the prior distribution given by the encoder. As recommended by \citep{kingma_auto-encoding_2014}, $L = 1$ is enough for training the model, but our experiments also find that a larger $L$ benefits especially for the uncertainty estimation. In Appendix  \ref{app:hyper}, an ablation study will be provided for the extra KL loss term and larger sampling size proposed in this section.

\subsubsection{Inference with GS-ED model}

Once the model parameters are learned, inference of the output $\bm y$ from input $\bm x$ and $c$ follows the generative process of the GS-ED model. The latent distribution $p_\theta(\bm z | \bm x) \sim \mathcal N \left(\bm \mu_E(\bm x^{(i)}, \bm \sigma_E(\bm x^{(i)})\right)$ is derived from the encoder. The posterior distribution of output $\bm y$ is then estimated using Monte Carlo sampling from this prior. Suppose $N_s$ latent variables $\left\{z^{(s)}\right\}_{s=1,\cdots,N_s}$ are samples, and $\hat {\bm y}^{(s)} = \bm \mu_D (\bm z^{(s)}, c)$ is the decoder output corresponding to the $s$-th sample, the mean and standard deviation of $p_\theta(\bm y | \bm z,c)$ can be estimated as followed:

\begin{equation}
     \mu(\bm{\hat y}) = \frac{1}{N_s}\sum_{s=1}^{N_s}\bm{\hat y}^{(s)}, \quad  \sigma(\bm{\hat y}) = \sqrt{\frac{1}{N_s-1}\sum_{s=1}^{N_s}\left(\bm{\hat y}^{(s)}- \mu(\bm{\hat y})\right)^2}
     \label{eqn:uq}
\end{equation}

In real applications, several postprocessing steps are still required before predicting performance metrics from the primary outputs of the model. To propagate uncertainty through these steps, Monte Carlo sampling is extended to the postprocessing procedures. Suppose the performance metric $Y$ is deterministically obtained through postprocessing $\mathcal B$ given the model’s primary outputs $\bm y$ under a series of operating conditions $c$:

\begin{equation}
    Y = \mathcal B\left(\left\{c_j, \bm y_j\right\}_{j=1, \cdots, N_j}\right)
    \label{eqn:pp}
\end{equation}

UQ for the performance metrics still begins by sampling the latent space. For each sampled latent vector $\bm z^{(s)}$, primary outputs are generated under the specified operating conditions using the decoder: $\bm {\hat y}_j^{(s)}=\mu_D(\bm z^{(s)}, c_j)$. Subsequently, the performance metrics $Y^{(s)}$ that corresponding to this sample point are computed with Equation \ref{eqn:pp}. Following a procedure similar to Equation \ref{eqn:uq}, its probability distribution is estimated by:

\begin{equation}
     \mu(Y) = \frac{1}{N_s}\sum_{s=1}^{N_s}Y^{(s)}, \quad  \sigma(Y) = \sqrt{\frac{1}{N_s-1}\sum_{s=1}^{N_s}\left(Y^{(s)}- \mu(Y)\right)^2}
\end{equation}

Since the confidence interval upper bound is the value to be used in optimization, it can be derived with Equation \ref{eqn:ub} from $\mu(Y)$ and $\sigma(Y)$.

\subsubsection{Uncertainty calibration}

As discussed in Sec. \ref{sec:objintro}, the model uncertainty can be viewed irreducible with respect to the optimization process. Thus, calibration is essential after model training, so the model can make more precise uncertainty estimation when offline deployed. For samples in the training and testing datasets, if the predicted confidence intervals are accurate, the actual performance derived from CFD-simulated flow fields should fall within the predicted confidence bounds with the expected frequency: $p(LB_\alpha\le Y^{(\mathrm{CFD})}\le UB_\alpha) \to \alpha$.

Calibration is carried out by linearly adjusting the $LB$ and $UB$ across all training samples. Specifically, calibration factors $\kappa_L$ and $\kappa_U$ are introduced to ensure that the proportion of errors falling below LB or above UB is $\frac{1}{2}(1-\alpha)$, respectively.

\begin{align}
    LB'_\alpha=\mu(Y) - \kappa_L \frac{t_{\alpha,N_{s-1}}}{\sqrt{N_s}}\cdot \sigma(Y),\quad UB'_\alpha=\mu(Y) + \kappa_U \frac{t_{\alpha,N_{s-1}}}{\sqrt{N_s}}\cdot \sigma(Y)\nonumber\\
    s.t. \quad p(Y^{(\mathrm{CFD})}\le LB'_\alpha) \to \frac{1}{2}(1-\alpha), \quad p(UB'_\alpha \le Y^{(\mathrm{CFD})}) \to \frac{1}{2}(1-\alpha)
    \label{eqn:uc}
\end{align}

\section{Experiment setup}\label{sec3}

\subsection{Shape optimization problem}

Multipoint aerodynamic shape optimization provides a good case for demonstrating the DBO and UA-DBO frameworks. For next-generation civil aircraft, maintaining high aerodynamic efficiency and robustness across off-design conditions is critical \citep{martins_aerodynamic_2022}. However, efficient multipoint aerodynamic optimization remains a major challenge. Traditional CFD-based frameworks require multiple simulations to evaluate multipoint performance, resulting in significant computational expense. In contrast, DBO frameworks enable performance prediction independent of the number of design points by leveraging a pretrained model, dramatically accelerating optimization \citep{li_data-driven_2019, li_physics-based_2022}. 

This study uses the optimization problem of supercritical airfoil's drag divergence performance to demonstrate the performance of the proposed UA-DBO method. As the Mach number $Ma_\infty$ or angle of attack $AOA$ increases, shock waves on the airfoil's upper surface intensify, causing a rapid rise in drag, which is referred to as drag divergence. 
The model training, uncertainty quantification, optimization process and results are comprehensively analyzed. 

The optimization problem is based on the second benchmark case of the aerodynamic design optimization discussion group (ADODG) \citep{martins_aerodynamic_2022}. The objective is to minimize the average drag $\bar C_D$ of the RAE2822 airfoil across six Mach numbers (0.710 to 0.760) while maintaining constraints on sectional area $A$ and cruise pitching moment $C_{M,\mathrm{cruise}}$. The lift coefficient $C_{L,\mathrm{cruise}}$ is fixed at 0.824 for all Mach numbers. To ensure engineering practicality, the airfoil thickness at 15\% chord $t_{0.15c}$ must not be less than that of the baseline airfoil. Airfoil shapes are parameterized using two sets of ninth-order class shape transformation (CST) parameters for the upper and lower surfaces, denoted as $\{u_i\}_{i = 0,\cdots, 9}$ and $\{l_i\}_{i = 0,\cdots, 9}$. The optimization problem can be mathematically expressed as follows:

\begin{align}
    \min_{\{u_i, l_i\}_{i = 0,\cdots, 9}} \bar C_D &= \frac{1}{6}\sum_{Ma_\infty=0.71, 0.72, \cdots, 0.76}C_{D, Ma_\infty}\\
    s.t. \quad & C_{L,\mathrm{cruise}} = 0.824,\nonumber\\
    &A\ge A^{(0)}, ~ C_{M,\mathrm{cruise}} \ge C_{M,\mathrm{cruise}}^{(0)},\nonumber\\
    & 1.0 \le AOA_{\mathrm{cruise}} \le 5.0\nonumber\\
    & t_{0.15c} > 0.9 \cdot t_{0.15c}^{(0)}\nonumber
\end{align}

\noindent where the values of the baseline airfoils are denoted with superscript $(\cdot)^{(0)}$.

To further demonstrate UA-DBO performance in multi-objective optimization, we test it on another optimization problem for buffet onset which is the same as our previous buffet onset optimization study \citep{yang_fast_2024}. Due to the limitation of space, we put the setup and results of this problem in Appendix \ref{app:buffet}.

\subsection{Model setup for the optimization problem}

\subsubsection{Prior-based strategy for multipoint aerodynamic performance prediction}

\begin{figure}[htbp]
    \centering
    \includegraphics[width=0.8\linewidth]{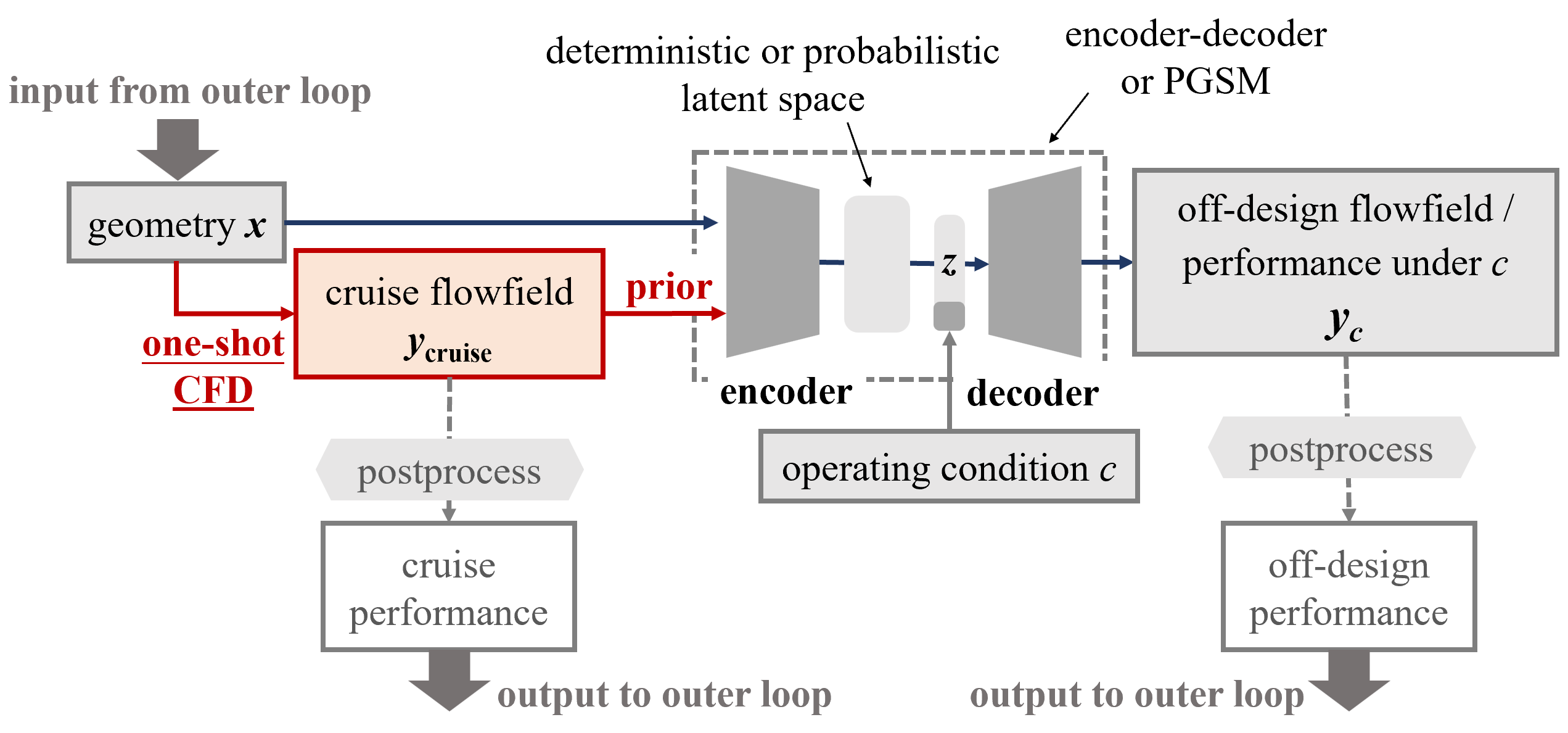}
    \caption{Prior-based framework for multipoint performance prediction}
    \label{fig:pb}
\end{figure}

In this work, the prior-based strategy proposed by \cite{yang_fast_2024} is employed to construct surrogate models for evaluating multipoint performance. A schematic of the prior-based surrogate model is shown in Fig. \ref{fig:pb}. Unlike direct prediction approaches that estimate off-design flow fields and performance metrics solely from airfoil geometry, the prior-based method incorporates a one-shot CFD simulation to calculate an accurate cruise flow field. The surrogate model is then trained to predict off-design performance based on both the airfoil geometry and the CFD-simulated cruise flow field. As the strategy only modifies the model inputs, it remains compatible with both deterministic and probabilistic data-based optimization frameworks. Specific model inputs, outputs, and architectures are detailed in Appendix \ref{app:model}. 

This prior-based approach mitigates the limitations of relying exclusively on pretrained models during optimization, thereby enhancing the model’s generalization capability and improving the robustness of DBO frameworks. Meanwhile, it preserves computational efficiency, enabling the evaluation of off-design performance within seconds and achieving multipoint optimization at a time cost comparable to single-point optimization.

\subsubsection{Model architecture}

\begin{figure}[htbp]
    \centering
    \includegraphics[width=0.7\linewidth]{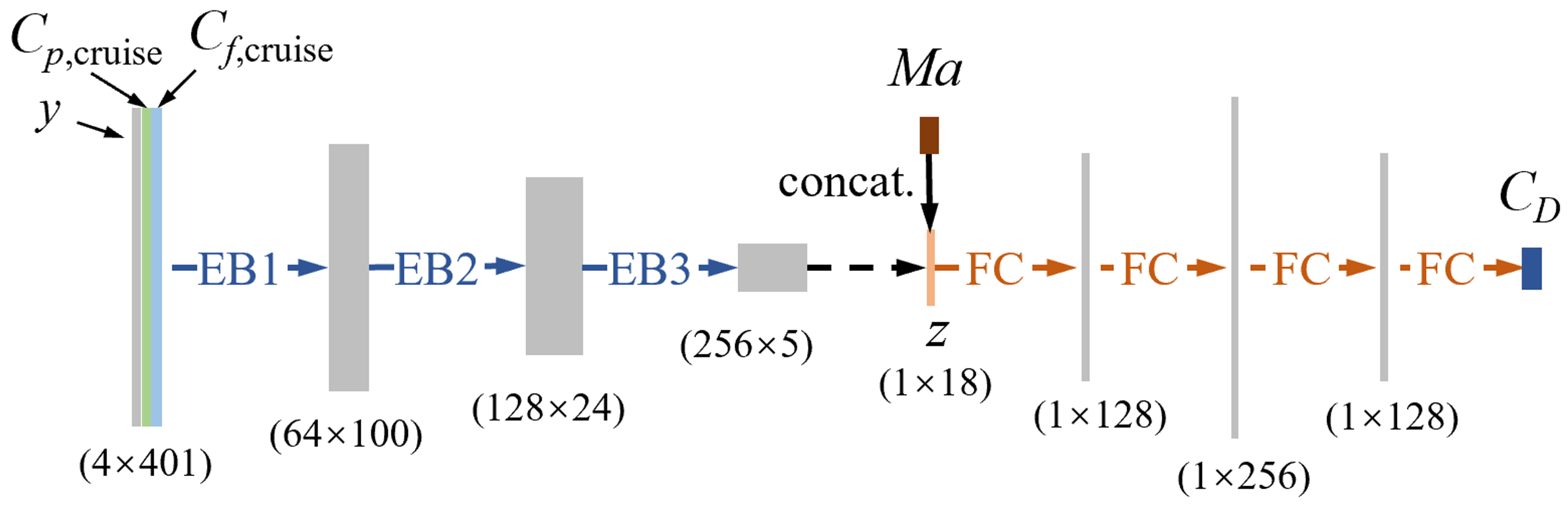}
    \caption{Architecture of the model for drag divergence optimization (The EB\# stands for the encoder block, and the FC stands for the fully connected layer. The values in brackets indicate the channel number and feature map sizes of the tensors.)}
    \label{fig:p1mod}
\end{figure}

The model architecture for drag prediction is shown in Fig. \ref{fig:p1mod}. It is designed to predict the drag coefficient for a given Mach number of an airfoil, enabling straightforward calculation of the average drag coefficient across a range of Mach numbers. In this problem, only the Mach number varies; all other operating conditions are held constant. 

The model inputs include the airfoil geometry, surface $C_p$ and $C_f$ distributions. As outlined in the previous section, the pressure and friction distributions on each surface are sampled at fixed $x$-positions. Here, we use the $y$-coordinates of the surfaces at these $x$-positions to describe the airfoil geometry. This ensures that all input vectors share the same shape and can be processed using an encoder based on 1D convolutional layers. After encoding, the Mach number corresponding to the target operating condition is concatenated to the latent vector $z$ of dimension 18, which is then decoded by a multi-layer perceptron (MLP) to generate the output.

The encoder comprises three blocks, each consisting of a 1D convolutional layer with a kernel size of 3 and a stride of 2, an average pooling layer, and a LeakyReLU activation layer with a slope of 0.2. The tensor sizes after each block are indicated in parentheses in Fig. \ref{fig:p1mod}, where the first value represents the number of channels and the second denotes the feature map length. Following the final encoder block, a fully connected layer links the flattened output to the latent vector $z$. In the probabilistic model, two fully connected layers predict the mean and log variance of $z$, respectively, and the reparameterization trick is used to sample $z$ by adding scaled white noise $\epsilon$.

The decoder contains three hidden layers with dimensions 128, 256, and 128, respectively, using batch normalization and LeakyReLU activation. A final fully connected output layer without activation predicts the drag coefficient.

\subsection{Pretraining settings of surrogate models}

\subsubsection{Airfoil database}

A airfoil dataset is built to pretrain the surrogate model by sampling the airfoil shape and operating conditions.

\begin{figure}[htbp]
    \centering
    \includegraphics[width=0.7\linewidth]{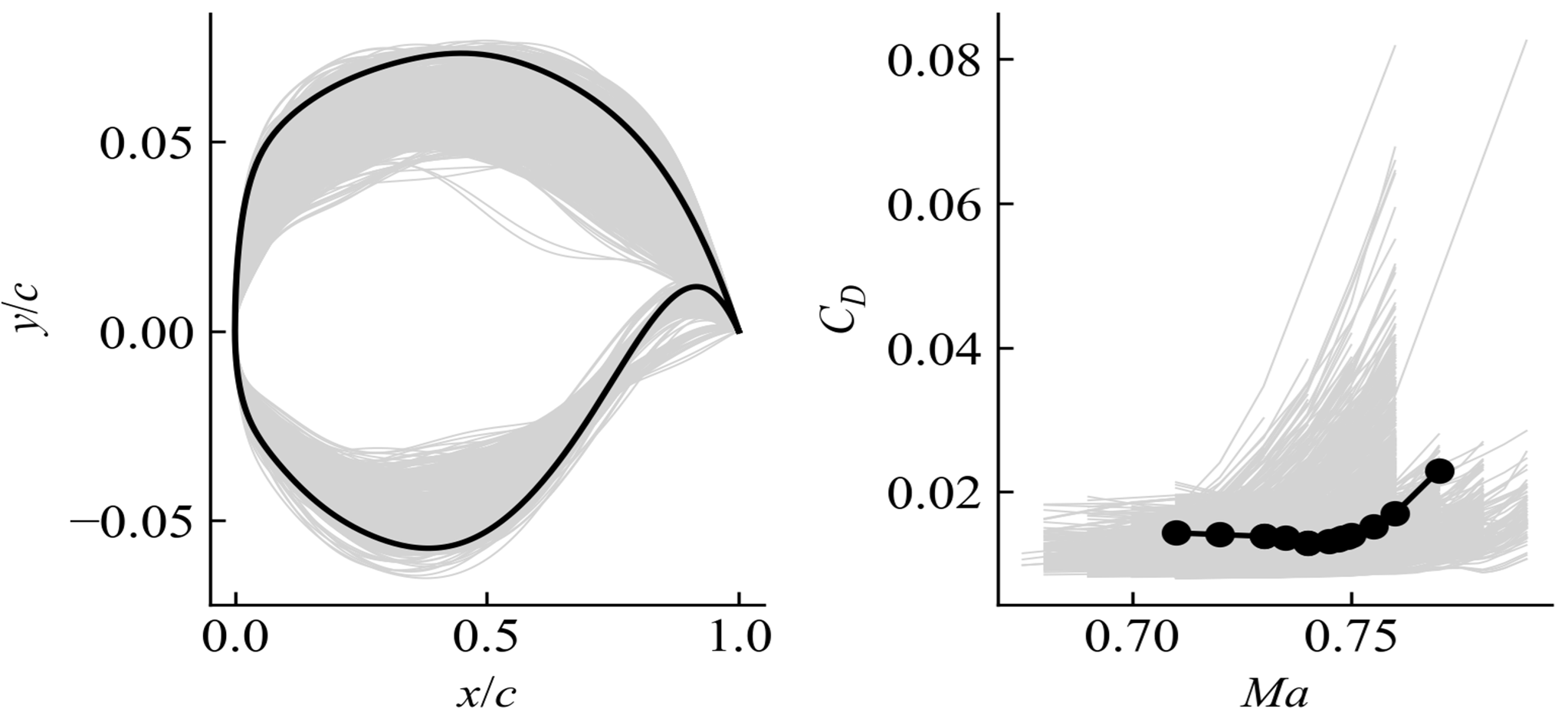}
    \caption*{
        \begin{minipage}{0.15\linewidth}
        ~
        \end{minipage}
        \begin{minipage}{0.3\linewidth}
            \centering
            \small (a) geometries of airfoils  
        \end{minipage}
        \begin{minipage}{0.08\linewidth}
        ~
        \end{minipage}
        \begin{minipage}{0.35\linewidth}
            \centering
            \small (b) drag divergence performance of airfoils in database
        \end{minipage}
    }
    \caption{Geometries and performance of airfoils in the databases}
    \label{fig:airfoils}
\end{figure}

\paragraph{Sampling of airfol shapes} Airfoil shapes and their cruise operating conditions are sampled using the output space sampling (OSS) method proposed by \cite{li_pressure_2022}, ensuring diverse patterns in geometry and pressure distribution. The total amount of airfoil shapes in the dataset is 1420, and the same airfoil set was used in our previous studies \citep{yang_fast_2024}. Fig. \ref{fig:airfoils}a shows these geometries, with a representative airfoil highlighted in black.

\paragraph{Sampling of operating condtions} After the airfoils and their cruise conditions are determined, off-design operating conditions are sampled according to each optimization task. Since off-design drag performance across multiple Mach numbers is needed, Mach numbers are randomly sampled within the range 0.65–0.80 for each airfoil, maintaining the same lift coefficient as in cruise. Fig. \ref{fig:airfoils}b shows the resulting $C_D - Ma_\infty$. 

\paragraph{CFD simulation}

The flow fields are computed using the open-source Reynolds-averaged Navier–Stokes solver CFL3D. Turbulence is modeled with the Shear Stress Transport (SST) model. A structured C-type grid, which is automatically generated using an in-house developed code, is employed for the simulations. To accelerate convergence, a W-cycle multigrid strategy is applied, with three grid levels executed sequentially with 1000, 1000, and 4000 iterations, respectively. The Reynolds number and freestream temperature are fixed at 20 million and 580 $^{\circ}$R, respectively. Only convergence CFD results are collected to construct the dataset, and the amounts of results is 17386.

Subsequently, the $C_p$ and $C_f$ distributions are extracted from simulated flow fields and normalized to a range of –1 to 1 based on their minimum and maximum values. The distributions are aligned by interpolating from the computational surface grid points to a set of specially designed $x$-positions, which are more densely distributed near the leading and trailing edges. The lift and drag coefficients $C_L, C_D$ can then obtained with $C_p$ and $C_f$.

Further details regarding database sampling and CFD simulations are available in our previous work \citep{yang_fast_2024} 

\subsubsection{Training settings}

Both the deterministic and probabilistic models in the DBO and UA-DBO frameworks are trained under identical settings. For deterministic models, the training objective is the mean squared error (MSE) between CFD-simulated and model-predicted results. For probabilistic models, the loss function is derived from Equation \ref{eqn:trainloss}. The weight for the extra KL loss term $\beta$ and the sampling size $N_l$ are determined through trial-and-error experiments to be $1\times10^{-5}$ and 4, respectively. During inference, the sampling size is 16.

During training, 82\% of the database flow fields are used for model training, with the remaining 18\% reserved for testing. Each training procedure is repeated three times for cross-validation. In each run, 10\% of the training samples are randomly selected for validation, and the model is trained on the remaining samples with randomized initial weights and biases. A batch size of 16 is used, and the Adam optimizer is adopted. A warmup strategy is applied, gradually increasing the learning rate from $1 \times 10^{-4}$ to $1 \times 10^{-3}$ over the first 20 epochs, followed by an exponential decay with a base of 0.95. Validation losses are continuously monitored to prevent overfitting, and the maximum number of training epochs is set to 300.

\subsection{Optimization settings}

\subsubsection{Baseline airfoils} 

\begin{figure}[htbp]
    \centering
    \includegraphics[width=0.8\linewidth]{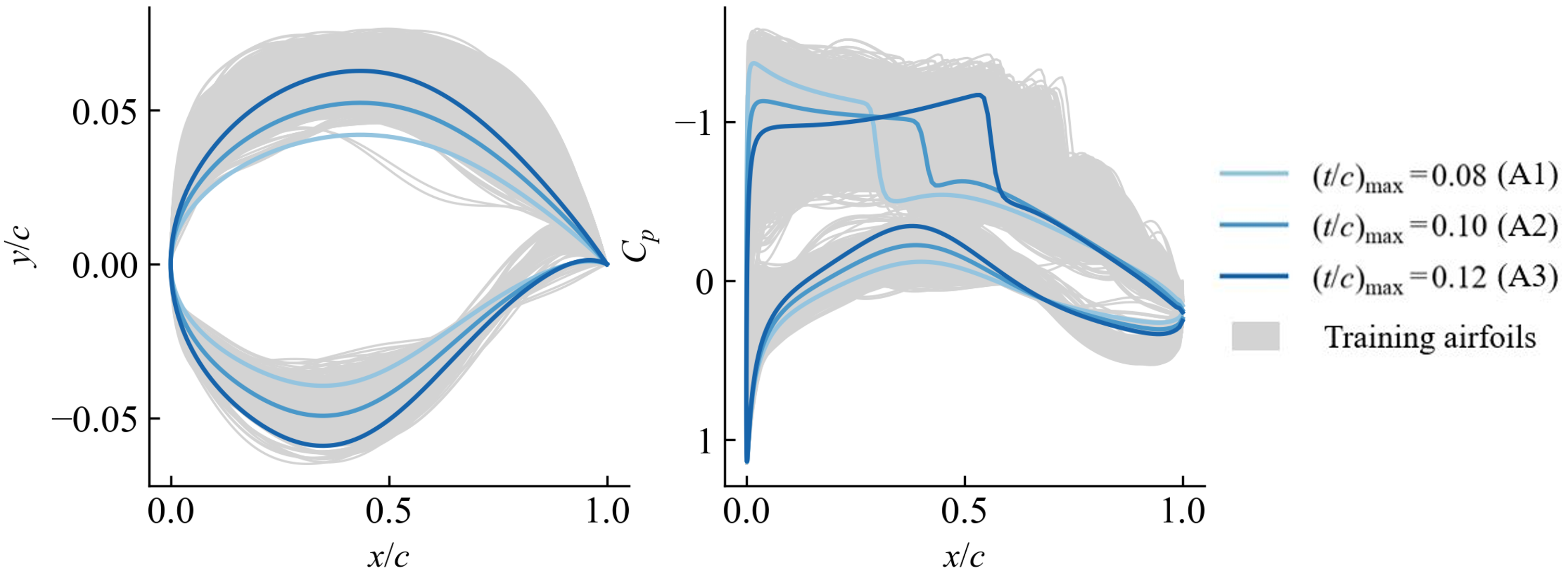}
    \caption{Geometries and cruise surface pressure distributions of baseline airfoils for drag divergence optimization}
    \label{fig:airfoilp1}
\end{figure}

Three baseline airfoils are used for demonstration: the RAE2822 airfoil and two variants obtained by scaling the RAE2822 airfoil in the y-direction to achieve maximum relative thicknesses of 0.081 and 0.101. The cruise operating condition for the prior-based model is set at $C_L = 0.824$ and $Ma_\infty = 0.730$. Fig. \ref{fig:airfoilp1} illustrates the baseline airfoil geometries and their corresponding cruise pressure coefficient Cp distributions. Baseline airfoils are shown in blue, with the training dataset samples overlaid in grey. Notably, the airfoil with a maximum relative thickness of 0.08 (Case A1) lies outside the training dataset.

\subsubsection{Optimization algorithm} 

The primary optimization algorithm employed is the multi-objective differential evolutionary algorithm, a global, gradient-free, and stochastic method that identifies optimal samples by iteratively updating a population through mutation, crossover, and selection based on individuals’ fitness values derived from their objective functions. The in-house code AeroOPT developed by \cite{li_pressure_2018} is used to implement this algorithm. All three optimization processes begin with an initial population of 64 airfoils, generated by adding random bumps to the baseline geometry and refitting the perturbed shapes using CST functions. Subsequently, 50 optimization iterations are performed, with a population size of 32 in each iteration.

\section{Results}

\subsection{Performance of pretrained surrogate models}\label{sec:perf}

Before being deployed for optimization, model performance on aerodynamic metric prediction is evaluated, and the advantages of the proposed GS-ED model are demonstrated. As mentioned earlier, we also provided a model detailed ablation study in the Appendix \ref{app:hyper}.

\subsubsection{Accuracy on aerodynamic coefficients}

In this section, we compare the proposed GS-ED model with the deterministic baseline in terms of prediction accuracy. The baseline model has the exact same Encoder and Decoder as the GS-ED model, except that its Encoder only outputs the deterministic latent value $z$, instead of predicting the mean and log standard error values of the latent Gaussian distribution. 

Table \ref{tab:errP1} reports the model prediction errors, where the first column presents the mean absolute errors (MAEs) between model-predicted and ground-truth drag coefficients across all training and testing samples, i.e., $\frac{1}{N_\mathrm{sample}}\sum_i^{N_\mathrm{sample}}\left|C_D^{\mathrm{(Model)}}-C_D^{\mathrm{(CFD)}}\right|$ where $N_\mathrm{sample}$ denotes the total number of samples. The second column shows the MAEs for the drag divergence performance metric used in optimization, defined by averaging errors among each airfoil $\frac{1}{N_\mathrm{airfoils}}\sum_i^{N_\mathrm{airfoils}}\left|\bar C_D^{\mathrm{(Model)}}-\bar C_D^{\mathrm{(CFD)}}\right|$.

\begin{table}[ht]
\centering
\caption{Prediction errors of models for $C_D$ prediction (in counts)}\label{tab:errP1}%
\begin{tabular}{@{}cccc@{}}
\toprule
Model & Dataset & MAE of $C_D$ for all samples & MAE of $\bar C_D$\\
\midrule
\multirow{2}{*}{deterministic ED}    & Training & 1.69 & 0.86 \\
                       & Testing  & 1.85 & 0.96 \\
\midrule
\multirow{2}{*}{GS-ED} & Training & 1.72 & 0.93 \\
                       & Testing  & 1.83 & 0.96 \\
\bottomrule
\end{tabular}
\end{table}

By comparing the errors of the deterministic and probabilistic models, it is found that the introduction of UQ slightly worsens the model's performance on the training dataset. This degradation arises from the complexity of the probabilistic loss term, which makes optimization more challenging, and from approximations inherent in modeling the true posterior distribution. However, the models perform similarly on the testing datasets, highlighting the superior generalization ability of the probabilistic model.

\subsubsection{Accuracy on uncertainty quantification}

This section further compares the UQ results of the pretrained model. Except for the proposed GS-ED, we also include a traditional non-penetrating UQ method, the deep ensemble method, to generate uncertainty metrics from the deterministic model. To be specific, the deep ensemble method separately trained $n$ deterministic models with different subsampling of 90\% of the training dataset and random seeds for initialization. During inference, these $n$ models are still seperately used to get predictions $\left\{\bm{\hat y}^{(i)}\right\}_{i=1,\cdots,n}$ and the ensembled predictions will be the mean $1/n\cdot \sum_{i=1,\cdots n}\bm{\hat y}^{(i)}$ and its uncertainty is obtained by $\sqrt{1/(n-1)\cdot \sum_{i=1,\cdots n}\left(\bm{\hat y}^{(i)} - \mu(\bm{\hat y})\right)}$.

For both deep ensemble and GS-ED, uncertainty calibration is performed to match the coverage of a 90\% confidence interval on the training dataset with Equation \ref{eqn:uc}. 
To demonstrate the UQ results, we tested the calibrated model on the unseen testing dataset with both the 90\% confidence interval converging and a more general expected calibration error (ECE) score. Here, we calculate the ECE based on the confidence interval. Nine confidence levels $\alpha = 0.1, \cdots 0.9$ are selected, and for each sample $(x_i, y_i)$, the model will give a confidence interval of $\alpha$, i.e., $\left\{\left[LB_\alpha(x_i), UB_\alpha(x_i)\right]\right\}_{\alpha= 0.1, \cdots,0.9}$. Let $
\hat c_\alpha = 1/N\sum_{i=1}^{N} \bm 1\left(y_i \in \left[LB_\alpha(x_i), UB_\alpha(x_i)\right]\right)$ where $\bm 1(\cdot)$ is the indicator function. Then $\hat c_\alpha$  indicates the fraction of true values that fall inside the predicted interval. The ECE is then defined by averaging the distance between $\hat c_\alpha$ and target $\alpha$: $ECE = \sum_{\alpha=0.1, \cdots 0.9} w_\alpha \left|\hat c_\alpha - \alpha\right|$ where $w_\alpha = 1/9$. 

Table \ref{tab:erruq} shows the model calibration results both on the training dataset (based on which the model is calibrated) and the testing dataset (which is unseen for the model). If the model is perfectly calibrated, the ground-truth values should fall within the predicted interval with the target confidence level. This means that for a 90\% confidence interval coverage, the percentage of samples below the lower bound, within the interval, and above the upper bound should be 5\%, 90\%, and 5\%, respectively. For ECE, the perfectly calibrated model corresponds to an ECE of 0. Here, we test the deep ensemble method with $n=3 ~\mathrm{or}~ 6$. For the GS-ED model, one-time training is enough to get both predictions and uncertainty values. However, for robustness, we cross-validated the results with three separate runs, following the same procedure as the previous tests, and provide the average performance in Table \ref{tab:erruq}. To better compare the methods, the training and inference costs are also included in the last two columns, which were obtained on NVIDIA P2000 GPU.

\begin{table}[ht]
\centering
\caption{Uncertainty quantification results after calibration of the model for $\bar C_D$ prediction}\label{tab:erruq}%
\begin{tabular}{@{}ccccccccc@{}}
\toprule
\multirow{2}{*}{Model} & \multirow{2}{*}{Dataset} & \multicolumn{3}{c}{Frequency of} & \multirow{2}{*}{ECE} & \multicolumn{2}{c}{GPU Time} \\ \cmidrule{3-5}\cmidrule{7-8}
 & & \makecell{$Y<$ \\ $LB_{90\%}$} & \makecell{$LB_{90\%} \le Y$ \\$\le UB_{90\%}$} & \makecell{$UB_{90\%}$ \\$ < Y$} & & Training & Inference\\
\midrule
\multirow{2}{*}{\makecell{Ensemble \\ ($n=3$)}}  & Train & 5.0\% & 90.0\% & 5.0\% & 0.0786 & \multirow{2}{*}{1.41 hr} & \multirow{2}{*}{0.27 sec}\\
& Test  & 4.6\% & 89.2\% & 6.2\% & 0.0599\\
\midrule
\multirow{2}{*}{\makecell{Ensemble \\ ($n=6$)}}  & Train & 5.0\% & 90.0\% & 5.0\% & 0.0354 & \multirow{2}{*}{2.82 hr} & \multirow{2}{*}{0.54 sec}\\
& Test  & 3.1\% & 87.6\% & 9.2\% & 0.0345\\
\midrule
\multirow{2}{*}{GS-ED} & Train & 5.0\% & 90.0\% & 5.0\% & 0.0528 & \multirow{2}{*}{0.53 hr} & \multirow{2}{*}{0.10 sec} \\
& Test  & 3.6\% & 89.5\% & 6.9\% & 0.0491\\
\bottomrule
\end{tabular}
\end{table}

Overall, GS-ED and deep ensembles demonstrate comparable capability in estimating predictive uncertainty, with all models achieving good calibration. Specifically, both methods produce interval coverages close to the target 90\%, and their ECE values remain small. The GS-ED model achieves the ECE between ensemble with $n=3$ and $n=6$, while providing more accurate coverage of the 90\% interval on the test set. The deep ensemble with $n=6$ achieves the lowest ECE; however, its 90\% interval convergence is not particularly accurate. 

In terms of computational efficiency, GS-ED offers a clear advantage. It requires training only a single model and performs inference in a single forward pass. Although it involves sampling steps and requires decoding multiple latent vectors, their computational cost is minimal. In contrast, the deep ensemble approach requires training multiple independent networks, increasing the training and inference time scaling proportionally. Although ensemble members can be trained in parallel on multiple GPUs, this approach still incurs additional hardware costs. Taken together, these results demonstrate that the proposed GS-ED approach provides an efficient and accurate alternative for uncertainty estimation.

In the following sections, the GS-ED model with the best 90\% interval coverage is used for optimization.


\subsection{Optimization results}

Three optimization frameworks are compared in this study: 
\begin{itemize}
    \item CFD-based framework: No surrogate model is employed. Instead, six CFD simulations are conducted to evaluate drag performance across Mach numbers ranging from 0.71 to 0.76.
    \item Original DBO framework: A deterministic encoder-decoder predicts the multipoint performance, and the optimization objective is solely based on the model-predicted performance metrics.
    \item Proposed UA-DBO framework: The GS-ED model predicts both the multipoint performance and its associated uncertainty, and a model-confidence-aware objective function guides the optimization.
\end{itemize}

The three optimization frameworks are evaluated on three test cases based on different baseline airfoils. To account for stochasticity and avoid causal artifacts, each model-based optimization (DBO and UA-DBO) is repeated five times with different initial populations and random seeds.

\paragraph{UA-DBO yields better and more stable actual optimization results}

Table \ref{tab:opt1} provides a quantitative comparison of the reductions in $\bar C_D$ achieved by the three frameworks. For the model-based approaches (DBO and UA-DBO), the first three columns summarize the performance of the model-optimal design: the model-predicted reduction value, its corresponding CFD-validated reduction value, and the prediction error between the two. For UA-DBO, the model-predicted values represent the expectation under its uncertainty-aware formulation. 

Due to model errors, the sample identified as optimal by the surrogate model may not correspond to the actual one, so all samples in the final population are subsequently evaluated using CFD to identify the actual best-performing individual discovered during model-based optimization. Its $\bar C_D$ reduction is reported in the last column and compared directly against the result of the CFD-based optimization. All reported values include mean and standard error across the five independent optimization runs.

\begin{table}[h]
\centering
\caption{Comparison of model-predicted and actual reductions of $\bar C_D$ (in counts)}\label{tab:opt1}%
\begin{tabular}{@{}cccccc@{}}
\toprule
\multirow{2}{*}{Case} & \multirow{2}{*}{Method} & \multicolumn{3}{c}{Model-based optimal} & \multirow{2}{*}{\makecell{Actual best in \\last population}}\\\cmidrule{3-5}
 &  & Model Pred. & CFD Eval. & Error & \\
\midrule
\multirow{3}{*}{A1\footnotemark[1]} & CFD-based & & & & 33.5\\
& DBO & 22.6 $\pm$ 11.4 & 12.7 $\pm$ 8.0 & 9.9 $\pm$ 9.3 & 15.3 $\pm$ 8.1  \\
& UA-DBO & 33.2 $\pm$ 8.7 & 26.9 $\pm$ 7.4 & 8.1 $\pm$ 3.8 & 28.9 $\pm$ 7.0 \\
\midrule
\multirow{3}{*}{A2}  & CFD-based & & & & 23.4\\
& DBO & 19.8 $\pm$ 5.3 & 16.0 $\pm$ 2.4 & 4.5 $\pm$ 2.7 & 17.9 $\pm$ 3.0  \\
& UA-DBO &20.9 $\pm$ 5.5 & 22.7 $\pm$ 4.6 & 3.3 $\pm$ 1.5 & 22.8 $\pm$ 4.7 \\
\midrule
\multirow{3}{*}{A3} & CFD-based & & & & 91.5\\
& DBO & 92.9 $\pm$ 11.0 & 65.4 $\pm$ 23.3 & 27.5 $\pm$ 14.6 & 69.9 $\pm$ 20.9 \\
& UA-DBO & 91.6 $\pm$ 3.9 & 87.5 $\pm$ 6.0 & 7.2 $\pm$ 6.0 & 87.7 $\pm$ 5.7 \\
\bottomrule
\end{tabular}
\footnotetext[1]{The maximum relative thickness of the baseline airfoil is beyond the training dataset}
\end{table}

\begin{figure}[htbp]
    \centering
    \includegraphics[width=0.95\linewidth]{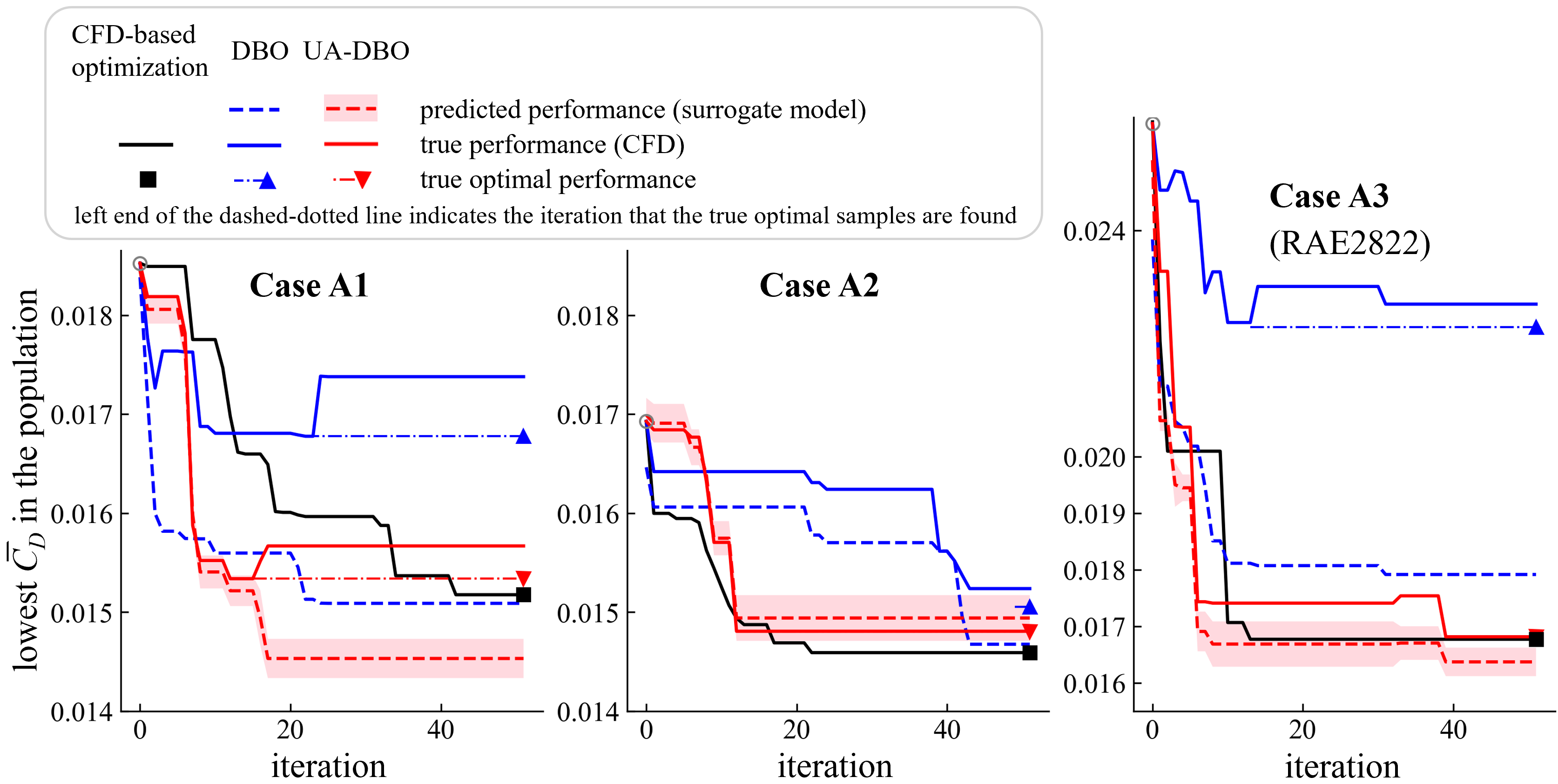}
    \caption{Optimization process of CFD and surrogate-based frameworks on drag divergence optimization}
    \label{fig:otp1}
\end{figure}

For each optimization case, we also select one run among five and show its convergence of the objective function during iterations in Fig. \ref{fig:otp1}. Starting from the $\bar C_D$ of baseline geometries (gray circles), the black lines trace the minimum $\bar C_D$ achieved by the CFD-based optimization. The blue and red dashed lines represent the best model-predicted "apparent” $\bar C_D$ minima obtained by DBO and UA-DBO, respectively. The pink shaded areas indicate the associated 90\% confidence intervals provided by the probabilistic model in UA-DBO. While DBO directly optimizes the model-predicted minima, UA-DBO optimizes the upper boundary of the confidence interval, corresponding to the upper edge of the pink area in the figure.

We also verify the optimization trajectories with CFD simulation, and actual $\bar C_D$ are obtained for the best sample of each iteration, allowing the actual optimization trajectories to be reconstructed and plotted as solid red and blue lines. The actual optimal in the last population (corresponding to the last column in Table \ref{tab:opt1})  are represented as black squares (CFD-based), blue triangles (DBO), and red deltas (UA-DBO), and the left ends of dashed-dotted lines in the figure indicate when these samples are discovered. It is worth noting that not all actual optimal samples are on the trajectory, since there may be samples with better model-predicted performance in the population.  

The results in Table \ref{tab:opt1} and Fig. \ref{fig:otp1} highlight the consistent advantages of the uncertainty-aware framework. In all cases, both DBO and UA-DBO yield airfoils that outperform the baseline, but DBO exhibits increasing prediction errors as optimization proceeds, ultimately resulting in false minima, where the predicted performance is better but the actual performance deteriorates. UA-DBO improves the actual optimization performance across all three cases by incorporating uncertainty information. Throughout the optimization, UA-DBO exhibits a smaller error in its performance, resulting in a more accurate drag reduction when the model-predicted reduction is similar for both UA-DBO and DBO frameworks. The gains are particularly good in Cases A1 and A3, where UA-DBO discovers the actual best sample with 13 $\sim$ 17 counts better than DBO. On average, UA-DBO delivers a drag reduction 1.47 times greater than that achieved by DBO and reduces the model error of the optimized sample by 39.6\%. 

Meanwhile, UA-DBO also substantially reduces the variance of prediction error in all cases, indicating that uncertainty-guided acquisition not only improves the reliability of model estimates but also stabilizes the optimization trajectory across repeated runs. 


\paragraph{Uncertainty information prevents overconfident model predictions}

\begin{figure}[H]
    \centering
    \includegraphics[width=1\linewidth]{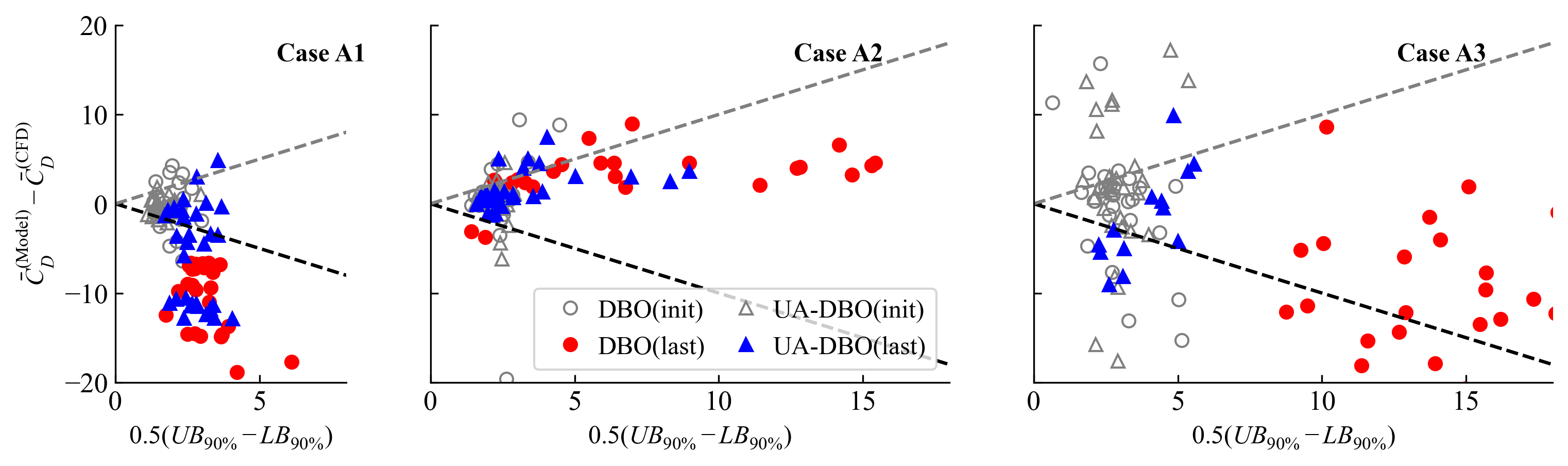}
    \caption{Model-predicted confidence interval and actual errors of the initial and final populations (axis values in counts)}\label{fig:errors}
\end{figure}

Below, we further analyze how the uncertainty estimation of the pretrained GS-ED model influences the optimization behavior. Using the same experimental run as in Fig. \ref{fig:otp1}, we evaluate the predictive uncertainty for every individual in both the initial and final populations produced by DBO and UA-DBO. For each individual, the GS-ED model provides a predicted mean value together with the corresponding confidence interval. Figure \ref{fig:errors} visualizes the resulting model errors against the predicted 90\% confidence interval width. The gray markers denote individuals in the initial population, while the red and blue markers represent the final populations obtained by DBO and UA-DBO, respectively. The two dashed lines indicate the ideal calibrated bounds, within which a perfectly calibrated model should place 90\% of the errors. Importantly, the black dashed line corresponds to the upper confidence bound used during optimization; samples whose true error exceeds this bound are underestimated by the model and may lead the optimizer toward misleading solutions.

Across the initial populations of all three cases, the model exhibits reasonably accurate uncertainty quantification, as most samples fall within their predicted 90\% confidence intervals. In contrast, the geometry-OOD Case A1 does not exhibit substantially larger errors. The reason is that our optimization framework incorporates the cruise flow field as part of the inputs, along with geometry, effectively reducing the difficulty of predicting geometry OOD samples. The relatively larger errors in Case A3, on the other hand, may result from a higher baseline drag coefficient. 

The error distributions in the final populations clearly demonstrate the advantage of incorporating uncertainty. In Case A3, the GS-ED model assigns substantially larger predictive intervals to DBO’s final individuals (red points), preventing UA-DBO from being led astray by these falsely under-predicted optima. Indeed, the UA-DBO results (blue points) exhibit both smaller confidence interval widths and lower actual errors, confirming that it successfully navigates toward more reliable solutions. 

Case A2 exhibits another pattern, as the errors for both DBO and UA-DBO are small, and the UQ is the most accurate. Most errors in Case A2 correspond to over-predictions, yet UA-DBO drives the population toward regions with small predicted uncertainty, resulting in only marginal improvement over DBO. For Case A1, the model is less well-calibrated, so both DBO and UA-DBO include some samples that fall into a misleading under-estimated region. Nevertheless, the uncertainty estimates still allow UA-DBO to retain a subset of low-error individuals, thereby contributing to the improvement in optimization performance.

Overall, these analyses highlight that UA-DBO effectively avoids overconfident model predictions and leverages uncertainty information to guide the search toward robust and reliable optima, particularly in cases involving distributional shift such as A3. 


\paragraph{Incorporating uncertainty does not limit the ability to discover the optimal}

\begin{figure}[htbp]
    \centering
    \begin{subfigure}{1\textwidth}
        \centering
         \includegraphics[width=0.9\linewidth]{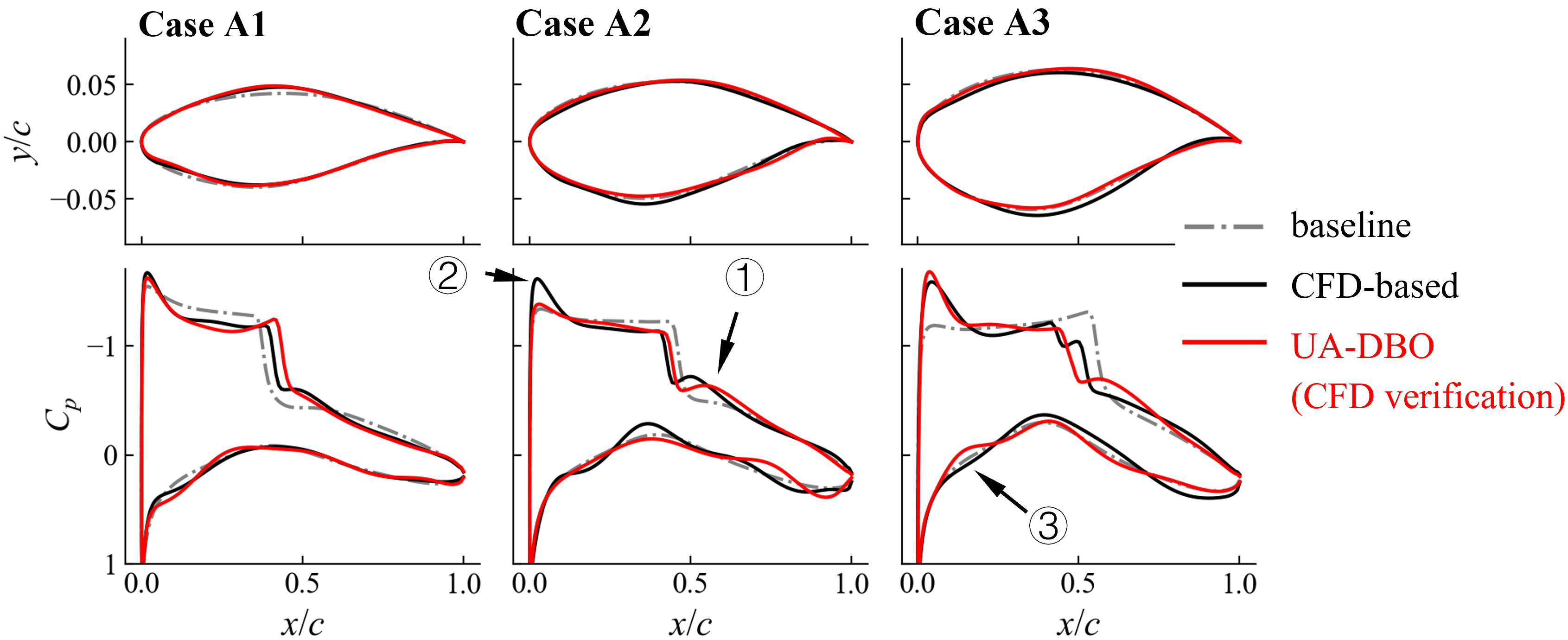}
        \caption{Airfoil geometries (upper) and cruise pressure coefficient distributions (lower)}
        \label{fig:p1surf}
    \end{subfigure}
    \hfill
    \begin{subfigure}{1\textwidth}
        \centering
        \includegraphics[width=0.75\linewidth]{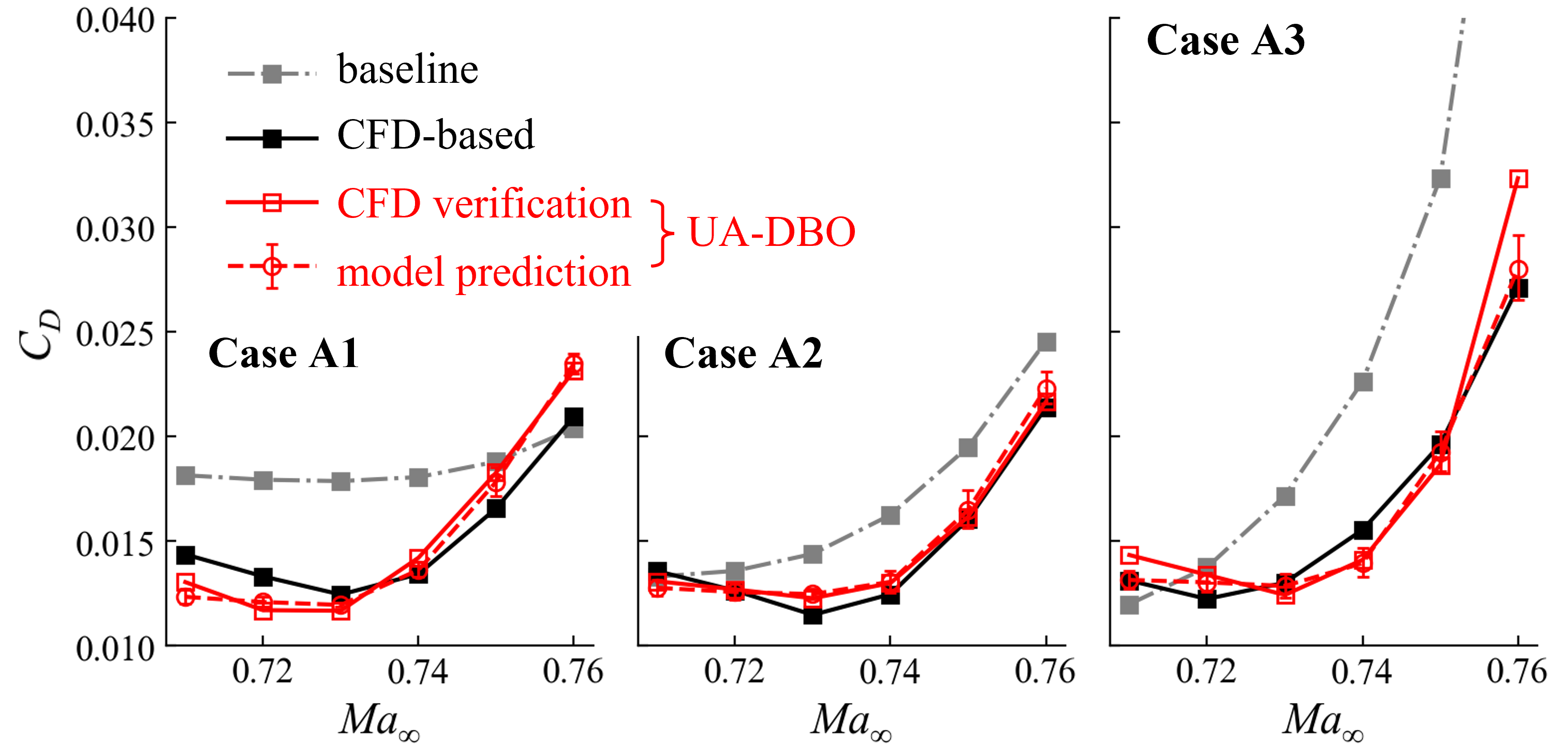}
        \caption{Drag coefficients as the Mach number increases}
        \label{fig:p1cd}
    \end{subfigure}
    \caption{Comparison of optimization results from CFD-based and UA-DBO}
    \label{fig:p1res}
\end{figure}

\begin{figure}[htbp]
    \centering
    \begin{subfigure}{1\textwidth}
        \centering
        \includegraphics[width=1\linewidth]{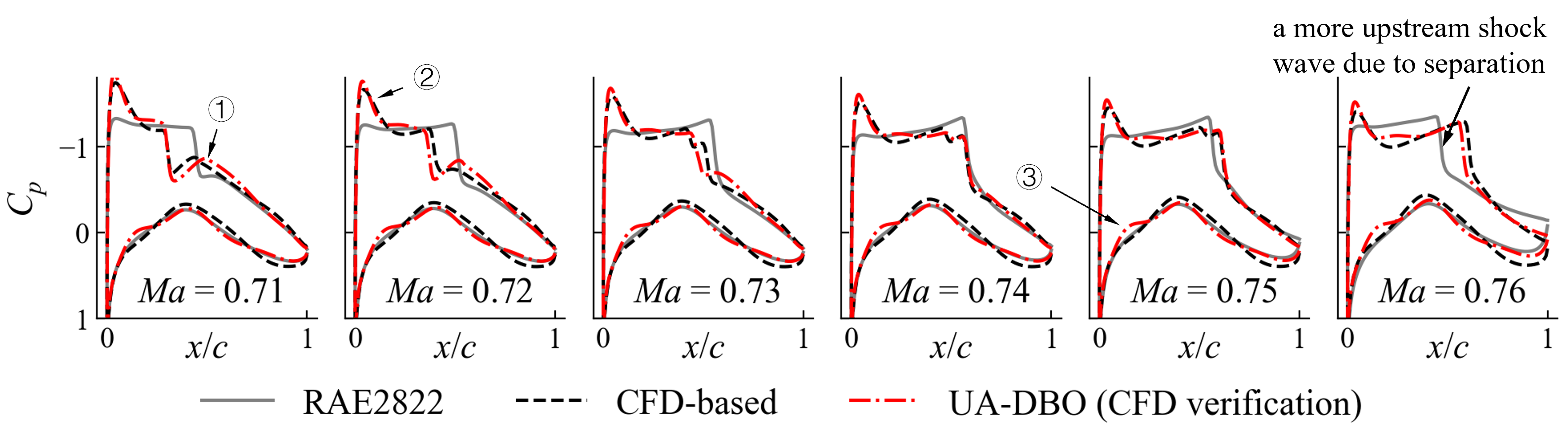}
        \caption{Surface pressure coefficient distributions}
        \label{fig:p1cp}
    \end{subfigure}
    \hfill
    \begin{subfigure}{1\textwidth}
        \centering
        \includegraphics[width=1\linewidth]{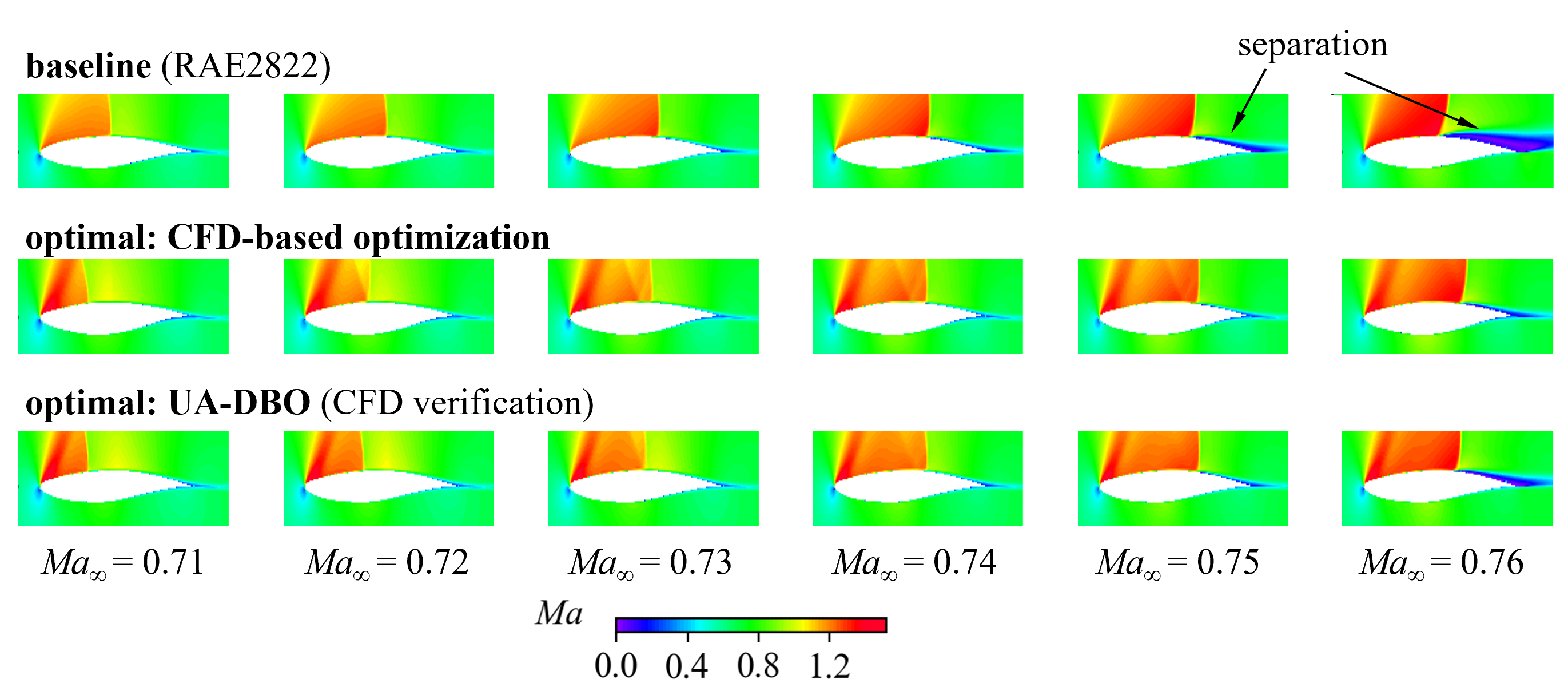}
        \caption{Mach number contours}
        \label{fig:p1ma}
    \end{subfigure}
    \caption{Comparison of optimization results from CFD-based and UA-DBO for the RAE2822 case (case A3)}
    \label{fig:p1field}
\end{figure}

We further comparing UA-DBO with CFD-based multipoint optimization. The results in Table \ref{tab:opt1} shows UA-DBO achieves an average reduction of 93.2\% in $\bar C_D$. A comparison of the best samples from UA-DBO and CFD-based optimization is presented in Fig. \ref{fig:p1res}, where \ref{fig:p1surf} illustrates the geometries and cruise pressure coefficients, and \ref{fig:p1cd} depicts the drag coefficient across Mach numbers. Fig. \ref{fig:p1field} shows the surface pressure distributions and the Mach number contours across Mach numbers for the baseline and optimized RAE2822 airfoils, where shock wave strength and separation can be intuitively observed.

Across all three cases, the CFD-based and UA-DBO exhibit a consistent trend of weakening the shock wave, as evident from the cruise pressure distributions. This confers a clear advantage for drag divergence performance by delaying shock wave intensification as Mach number increases and eliminating flow separation at high Mach numbers. 

On the other hand, a weaker shock wave leads to changes in other aerodynamic features that maintain lift, as indicated by the cruise pressure distributions. These include: (i) flow acceleration following the shock on the upper surface; (ii) an intensified suction peak near the leading edge; and (iii) the formation of a shallow cavity at the lower surface near the leading edge, known as front-loading, resulting in flow deceleration. These distinct mechanisms produce different local minima of drag coefficients and help explain the performance differences between optimization frameworks.

These results demonstrate that incorporating uncertainty does not limit the search region and hinders the optimization process from finding the optimal.

\subsection{Computation cost}

Table \ref{tab:time} summarizes the computational time required for model establishment and optimization for CFD-based, DBO, and UA-DBO frameworks. Both the detailed and a total wall-clock time for a complete optimization case and the total number of CFD evaluations are reported. All CFD simulations were performed using 8 Intel Xeon Gold 6145 CPU cores, while machine learning training and inference (underlined in the table) were executed on an NVIDIA P2000 GPU.
 
Consistent with our previous study \citep{yang_fast_2024}, DBO demonstrates a substantial efficiency advantage over conventional CFD-based optimization, even when the cost of database construction is taken into account. In the CFD-based framework, each optimization iteration requires multipoint CFD evaluations, leading to a total of 9,984 CFD calls and a wall-clock time of 41.6 hours for a single optimization case. In the buffet onset optimization provided in Appendix \ref{app:buffet}, more off-design points are need to evaluate the buffet performance, leads to even larger optimization time near 110 hours. Such a computational burden prevents real application, especially considering that optimization usually needs to be repeated several times. In contrast, both DBO and UA-DBO investigate most of the computational force in the offline model training phase. Once trained, the optimization can be finished with dramatically less time, which is almost equivalent to that of a single-point optimization, as the machine learning model’s inference time is negligible compared to a single CFD evaluation under cruise conditions.

\begin{table}[ht]
\caption{Computational Efficiency comparison}\label{tab:time}%
\begin{tabular*}{\textwidth}{@{\extracolsep\fill}cccc}
\toprule
  & CFD-based & DBO & UA-DBO \\
\midrule
data preparation & / & \multicolumn{2}{c}{830 hrs}  \\
model training & / & \underline{0.47 hrs} & \underline{0.53 hrs}  \\
\makecell{cruise evaluation  (one sample\footnotemark[1])} & \multicolumn{3}{c}{0.03 hrs} \\
\makecell{off-design evaluation  (one sample)} & 0.20 hrs & \underline{0.09 sec} & \underline{0.10 sec} \\
verify optimal samples\footnotemark[2] & / & \multicolumn{2}{c}{0.83 hrs}  \\
\midrule
\makecell{total number of CFD calls  (one case)\footnotemark[3]} & 9,984 & \multicolumn{2}{c}{1,696} \\
\makecell{total optimization time  (one case)\footnotemark[3]} & 41.6 hrs & \multicolumn{2}{c}{7.7 hrs} \\
\bottomrule
\end{tabular*}
\footnotetext[1]{performance evaluation for each sample is counted here with one processor, and the parallelization is between samples in each generation. }
\footnotetext[2]{samples in the last population of DBO and UA-DBO are verified with CFD to obtain the actual performance}
\footnotetext[3]{this shows the wall time for one optimization case run on 8 processors, with an initial population of 64, and later 50 iterations have a population of 32.}
\end{table}

The additional computational cost introduced by the proposed UA-DBO framework primarily arises from training the probabilistic model and quantifying uncertainty for each model prediction. Compared to deterministic models, the probabilistic approach includes reparameterization and sampling steps. It slightly prolongs the training time, and in experiments by 13\%. During model inference in optimization, UQ results in additional computation when $N_s$ latent vectors are sampled and passed to the decoder to generate the predicted flow field. Nevertheless, the decoding and postprocess procedures can be well paralleled, and theoretically it will only multiply the memory cost by $N_s$ while leaving training time largely unaffected when $N_s$ is small. Therefore, the efficiency of DBO is maintained in UA-DBO.

\section{Limitations and future work}

Despite the demonstrated effectiveness of the proposed UA-DBO framework, several limitations remain and point to directions for future research. Although the generalization capability of the surrogate model is enhanced through the inclusion of cruise flow fields into the input, and by the probabilistic nature of the GS-ED formulation, the proposed GS-ED model does not consistently maintain strong generalization performance when encountering geometries produced by optimization that deviate greatly from the training distribution. This issue is particularly noticeable for parameter combinations that are sparsely represented in the training dataset. In the optimization cases studied in this work, this limitation did not significantly compromise the effectiveness of UA-DBO; however, it indicates that further improvements are possible. 

Future work may explore augmenting purely data-driven confidence estimation with physics-embedded measures, such as uncertainty indicators derived from flow-field residuals or physical consistency constraints, to further improve the robustness and generalization of confidence prediction for unseen geometries.

\section{Conclusion}\label{sec13}

While DBO is a promising approach for rapid aerodynamic optimization, it lacks an inherent mechanism to quantify uncertainty in model predictions that are used during optimization. To address this limitation, this study introduces the UA-DBO framework, designed to enhance both the performance and reliability of DBO. The main contributions of this study are summarized as follows:

\begin{enumerate}
    \item A model-confidence-aware objective function for data-based optimization is proposed. Take minimizing the problem as an example. Unlike the original objective function which directly uses the performance metric to be optimized, UA-DBO uses a probabilistic model to predict the confidence interval upper bound of its prediction and uses it as the objective. It can simultaneously seek to maximize performance and minimize model uncertainty at the optimum, thereby avoiding the risk of optimization being misled by false optimal samples.
    \item To predict uncertainty with minimal modification to the original DBO framework, the GS-ED model is developed, adopting the variation approximation technique to quantify surrogate model uncertainty during offline deployment. We further proposed adding extra KL loss terms and increasing the sampling size to enhance its performance in uncertainty quantification. Compared to its deterministic counterpart, the GS-ED model maintains similar prediction accuracy, confirming that the introduction of uncertainty quantification does not compromise model performance. The calibrated GS-ED model is also compared with the deep ensemble method on its uncertainty estimation results, where GS-ED achieves an ECE of 0.049, similar accuracy, while reducing training cost by 1/5. 
    \item The DBO and UA-DBO frameworks are applied to the drag divergence optimization task. Compared to DBO, UA-DBO achieves a 1.47 times greater average drag reduction than DBO. When compared with CFD-based multipoint optimization, UA-DBO achieves 93.2\% of the average drag reduction, while maintaining a computational cost comparable to that of single-point optimization. We also demonstrate the performance of UA-DBO on a more complex multi-objective buffet onset problem, where it also produces superior and more reliable actual Pareto fronts than DBO.
    \item The relationship between the model’s predicted uncertainty and the corresponding actual prediction errors is further analyzed. The GS-ED model employed in UA-DBO predicts high uncertainty for the candidate solutions favored by DBO, which turn out to exhibit larger actual errors in most cases. By explicitly penalizing such high-uncertainty predictions during optimization, UA-DBO avoids being misled by falsely optimistic solutions.
\end{enumerate}

This study demonstrates that the efficiency and reliability of DBO can be significantly enhanced through the introduction of uncertainty quantification, paving the way for the broader application of data-driven models in aerodynamic optimization for engineering use.

\section*{Acknowledgements}

The authors would like to thank Weishao Tang from Tsinghua University, Assis. Prof. Lu Lu, Pengpeng Xiao, and Rui Chen from Yale University, for their inspiring comments. During writing, AI is used for polish some parts of this work. 

\section*{Declarations}

\paragraph{Fundings} This work was supported by the National Natural Science Foundation of China (NSFC) Nos. 12202243, 12372288, 12388101, and U23A2069. 

\paragraph{Conflict of interest} The authors have no Conflict of interest to declare that are relevant to the content of this article.

\paragraph{Author contributions} \textbf{Yunjia Yang}: Conceptualization (equal); Data curation (lead); Methodology (lead); Visualization (lead); Writing—original draft (lead). \textbf{Runze Li}: Conceptualization (equal); Data curation (supporting); Writing—review \& editing (equal). \textbf{Yufei Zhang}: Funding acquisition (supporting); Writing—review \& editing (equal). \textbf{Haixin Chen}: Funding acquisition (lead); Supervision (equal); Writing—review \& editing (equal).

\paragraph{Ethics approval and Consent to participate} Not applicable.

\paragraph{Data Availability} The datasets for model training are available upon reasonable request to the author, please see more information on \href{https://flogen.readthedocs.io/en/latest/airfoildataset.html}{this website}. 

\paragraph{Replication of Results} The codes for model training and optimization are available online with an open-source license at \href{https://github.com/YangYunjia/floGen}{GitHub}.

\begin{appendices}
\setcounter{figure}{0}

\section{Derivations of the GS-ED model}\label{secA1}

\begin{figure}[htbp]
    \centering

    \begin{subfigure}{\textwidth}
        \centering
        \includegraphics[width=0.5\linewidth]{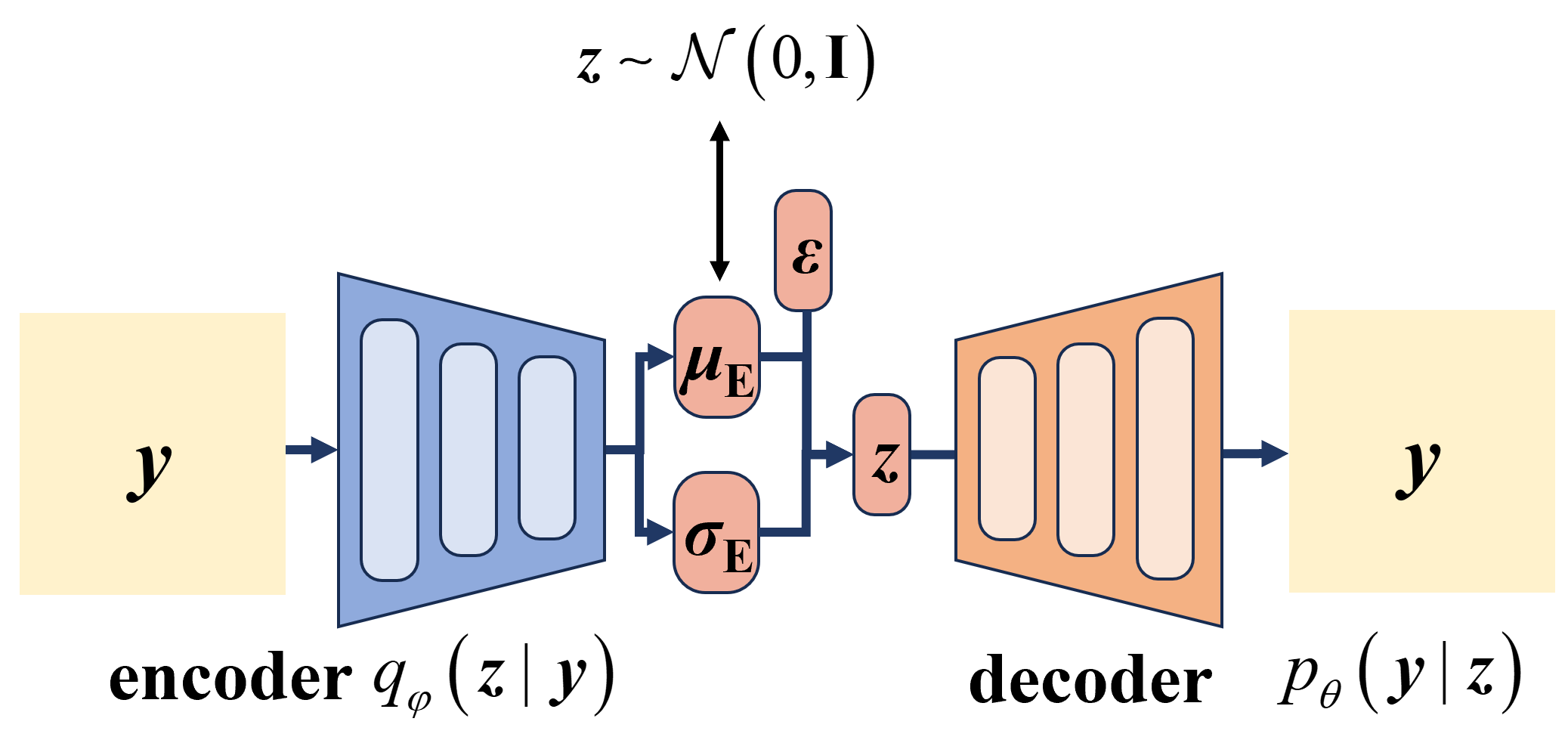}
        \caption{Variational auto-encoder (VAE)}
        \label{fig:vae}
    \end{subfigure}

    \begin{subfigure}{\textwidth}
        \centering
        \includegraphics[width=0.55\linewidth]{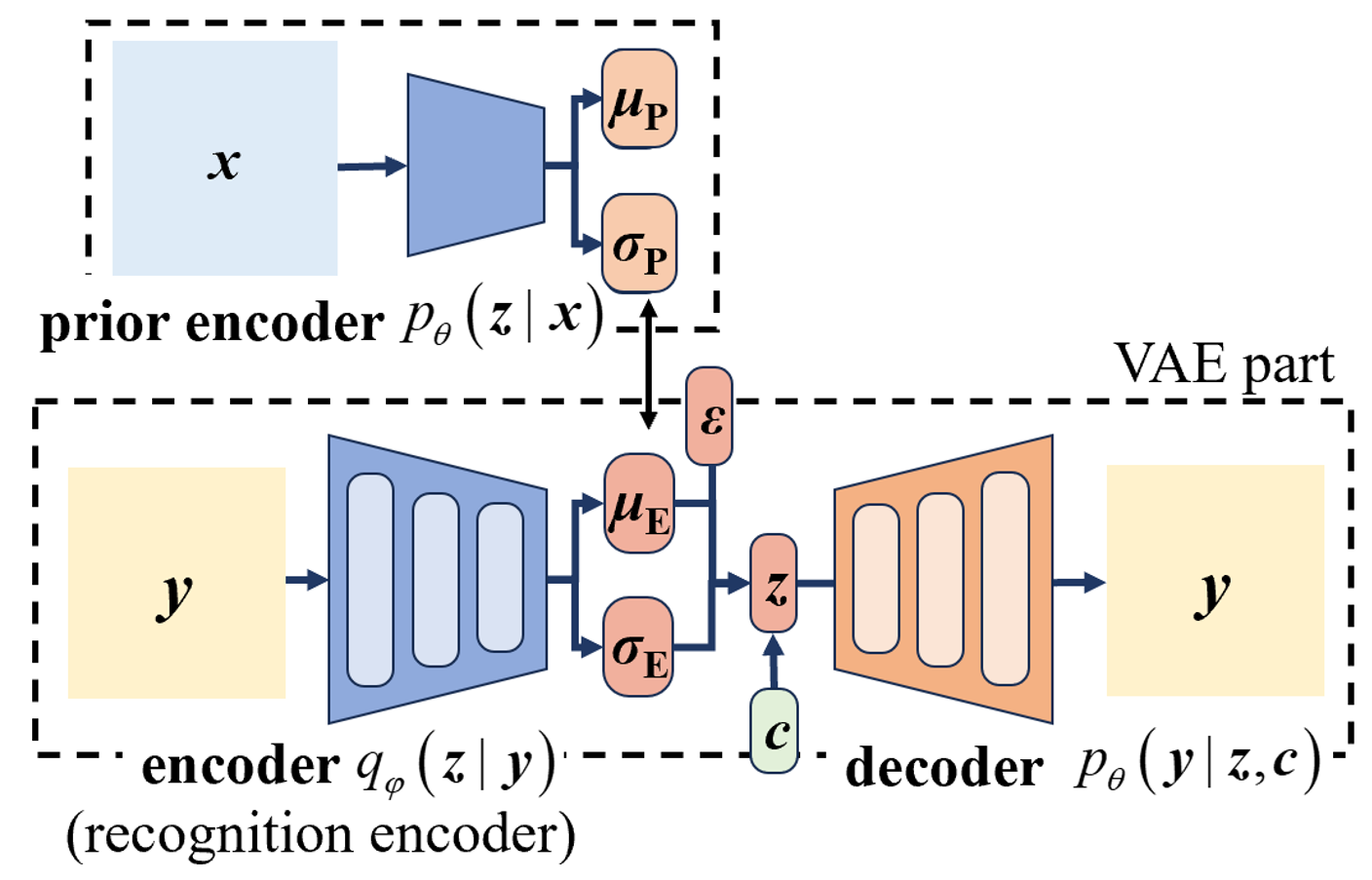}
        \caption{Prior-based variational auto-encoder (PVAE)}
        \label{fig:pvae}
    \end{subfigure}

    \begin{subfigure}{\textwidth}
        \centering
        \includegraphics[width=0.55\linewidth]{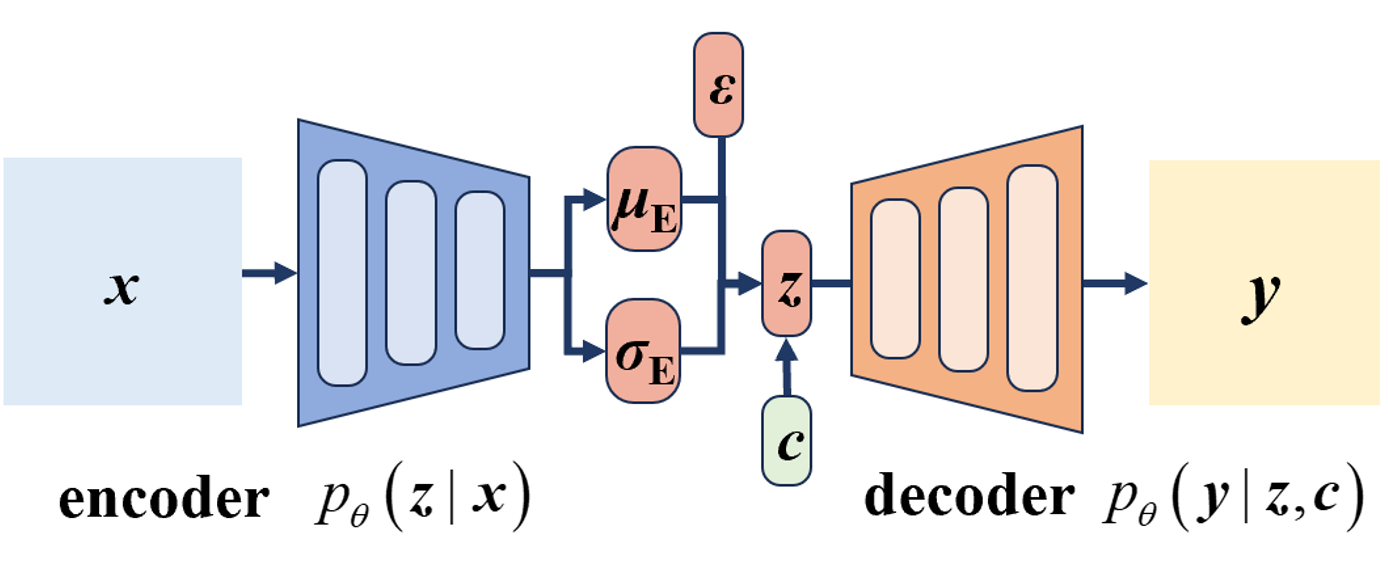}
        \caption{Gaussian stochastic encoder decoder (GS-ED)}
        \label{fig:pgsm1}
    \end{subfigure}
    \caption{Architectures of VAE, PVAE, and GS-ED models}
    \label{fig:mlmodels1}
\end{figure}

Since VAE is designed for dimension-reduction tasks, which means its input and output are the same, further development are made in this paper to make it fit the predictive tasks that infers flow field and performance given geometry and operating conditions. For this purpose, the prior-based variational autoencoder (PVAE) that was proposed by \cite{yang_flowfield_2022} is first introduced, and the GS-ED model is built by simplify its training process. A comparison of the VAE, PVAE, and GS-ED models are illustrated in Fig. \ref{fig:mlmodels1}

\subsection{Basics on VAE}

A typical VAE is designed for dimension reduction tasks. As shown in Fig. \ref{fig:vae}, its encoder functions as a recognizer that approximates the posterior distribution $p_\theta(\bm z | \bm y)$ of the latent variable $\bm z$ given the input $\bm y$, while the decoder serves as a generator that produces plausible samples from $p_\theta(\bm y | \bm z)$. To train the model, the variational approximation method is used, which introduces an approximate distribution $q_\phi(\bm z | \bm y)$ and tries to minimize the Kullback-Leibler (KL) divergence between the approximate and true distribution. Considering the evidence log-likelihood of the data $p_\theta(\bm y)$ also needs to be maximized, the evidence lower bound (ELBO) can be derived as:

\begin{align}
    ELBO_\mathrm{VAE} &= \log p_\theta(\bm y) - KL\left(q_\phi(\bm z | \bm y)||p_\theta(\bm z | \bm y)\right)
\end{align}

\noindent which needs to be maximized. By derivation, this leads to the training loss function of VAE as:

\begin{align}
    \min_{\theta, \phi}\mathcal L_\mathrm{VAE}(\bm y;\theta,\phi)\nonumber\\ \mathcal L_\mathrm{VAE}(\bm y;\theta,\phi)&=-ELBO_\mathrm{VAE}\nonumber\\&=-\mathbb E_{z\sim q_\phi(\bm z | \bm y)}\log p_\theta(\bm y|\bm z) + KL\left(q_\phi(\bm z | \bm y)||p(\bm z)\right)
\end{align}

Practically, the recognition distribution $q_\phi(\bm z | \bm y)$ is set to be a Gaussian distribution which is parameterized by its mean vector $\mu_E(\bm y^{(i)})$ and diagonal correlation matrix $\sigma_E(\bm y^{(i)})$; while the prior distribution $p(z)$ is a standard Gaussian distribution. With the definition of KL divergence, and reparameterization trick, the loss term can be derived to be

\begin{align}
    \mathcal L_\mathrm{VAE}(\bm y^{(i)};\theta,\phi)=\frac{1}{N_l}\sum_{l=1}^{N_l}\frac{1}{2}\left\Vert\bm y^{(i)}-\bm \mu_D(\bm z^{(i,l)})\right\Vert_2^2 \nonumber \\ +KL\left(\mathcal N\left(\mu_E(\bm y^{(i)}),\sigma_E(\bm y^{(i)})\right) \Vert \mathcal N (0, I)\right)
\end{align}

\noindent where $\bm z^{(i,l)}$ is sampled from $\mathcal N \left(\bm \mu_E(\bm y^{(i)}, \bm \sigma_E(\bm y^{(i)})\right)$.  

\subsection{PVAE and GS-ED}\label{subsubsec2}

To adapt VAE for flow field prediction task, the PVAE model is proposed in our previous work \citep{yang_flowfield_2022}, whose architecture is shown in Fig. \ref{fig:pvae}. PVAE aims to maximize the conditional evidence likelihood $\log p_\theta(\bm y | \bm x, c)$, rather than the plain one $\log p_\theta(\bm y)$. By similar derivation, the ELBO will be:

\begin{equation}
    -ELBO_\mathrm{PVAE}=-\mathbb E_{z\sim q_\phi(\bm z | \bm y)}\log p_\theta(\bm y|\bm z, c) + KL\left(q_\phi(\bm z | \bm y)||p(\bm z|\bm x)\right)
    \label{eqn:pvae}
\end{equation}

Compared to VAE, the prior distribution $p(\bm z)$ is conditioned by the input geometry $\bm x$, and the decoder is also designed to generate output $\bm y$ from both the latent $\bm z$ and the operating condition $c$. To obtain the conditioned prior distribution $p_\theta(\bm z | \bm x)$, a prior encoder, sharing the same architecture and parameters as the recognition encoder, is introduced. The output distribution is also a Gaussian and can be parameterized with $\bm \mu_P(\bm x^{(i)})$ and $\bm \sigma_P(\bm x^{(i)})$. A corresponding loss function for the PVAE is derived analogously

\begin{align}
    \mathcal L_\mathrm{PVAE}(\bm y^{(i)};\theta,\phi)=\frac{1}{N_l}\sum_{l=1}^{N_l}\frac{1}{2}\left\Vert\bm y^{(i)}-\bm \mu_D(\bm z^{(i,l)}, c^{(i)})\right\Vert_2^2 \nonumber \\ +KL\left(\mathcal N\left(\mu_E(\bm y^{(i)}),\sigma_E(\bm y^{(i)})\right) \Vert \mathcal N \left(\bm \mu_P(\bm x^{(i)}), \bm \sigma_P(\bm x^{(i)}\right)\right)
\end{align}

\noindent where $\bm z^{(i,l)}$ is still sampled from $\mathcal N \left(\bm \mu_E(\bm y^{(i)}, \bm \sigma_E(\bm y^{(i)})\right)$. The details about the derivation are provided in \cite{yang_flowfield_2022}.  

Training of PVAE involves iteratively updates the conditioned prior distribution $p_\theta(\bm z | \bm x)$ and the trainable parameters $\theta, \phi$ in the model, which leads to computation burden. GS-ED model simplifies the training process of PVAE inspired by \cite{sohn_learning_2015}. The prior encoder and the recognition encoder are set to be identical, thereby unifying the training and inference data flow (Fig. \ref{fig:pgsm1}). As a result, the second KL divergence term in Equation \ref{eqn:pvae} is no longer necessary, leading to the simplified ELBO:

\begin{equation}
    -ELBO_\mathrm{GS-ED}=-\mathbb E_{z\sim q_\phi(\bm z | \bm y)}\log p_\theta(\bm y|\bm z, c) 
\end{equation}
and the simplified loss function:

\begin{align}
    \mathcal L_\mathrm{GS-ED}(\bm y^{(i)};\theta,\phi)=\frac{1}{N_l}\sum_{l=1}^{N_l}\frac{1}{2}\left\Vert\bm y^{(i)}-\bm \mu_D(\bm z^{(i,l)}, c^{(i)})\right\Vert_2^2
\end{align}

\noindent where $\bm z^{(i,l)}$ is still sampled from $\mathcal N \left(\bm \mu_E(\bm x^{(i)}, \bm \sigma_E(\bm x^{(i)})\right)$ denotes the prior distribution given by the single encoder.

\section{Pre-experiments for hyperparameters}\label{app:hyper}
\setcounter{figure}{0}

The introduction of uncertainty requires two additional hyperparameters during model training: the weight of the extra KLloss term, $\beta$, and the sampling size $N_l$. We investigate how these hyperparameters influence both the predictive accuracy of the mean quantity and the quality of uncertainty estimation. Specifically, we evaluate four values of $\beta$, including the case without the extra KL term (denoted as $\beta = 0$) and three nonzero weights of $10^{-6}$, $10^{-5}$, and $10^{-4}$. For the sampling size $N_l$, we examine the commonly used choice of $N_l = 1$, as suggested in previous studies, as well as two larger values, $N_l = 4$ and $N_l = 8$.

\begin{figure}[htbp]
    \centering
        \includegraphics[width=0.8\linewidth]{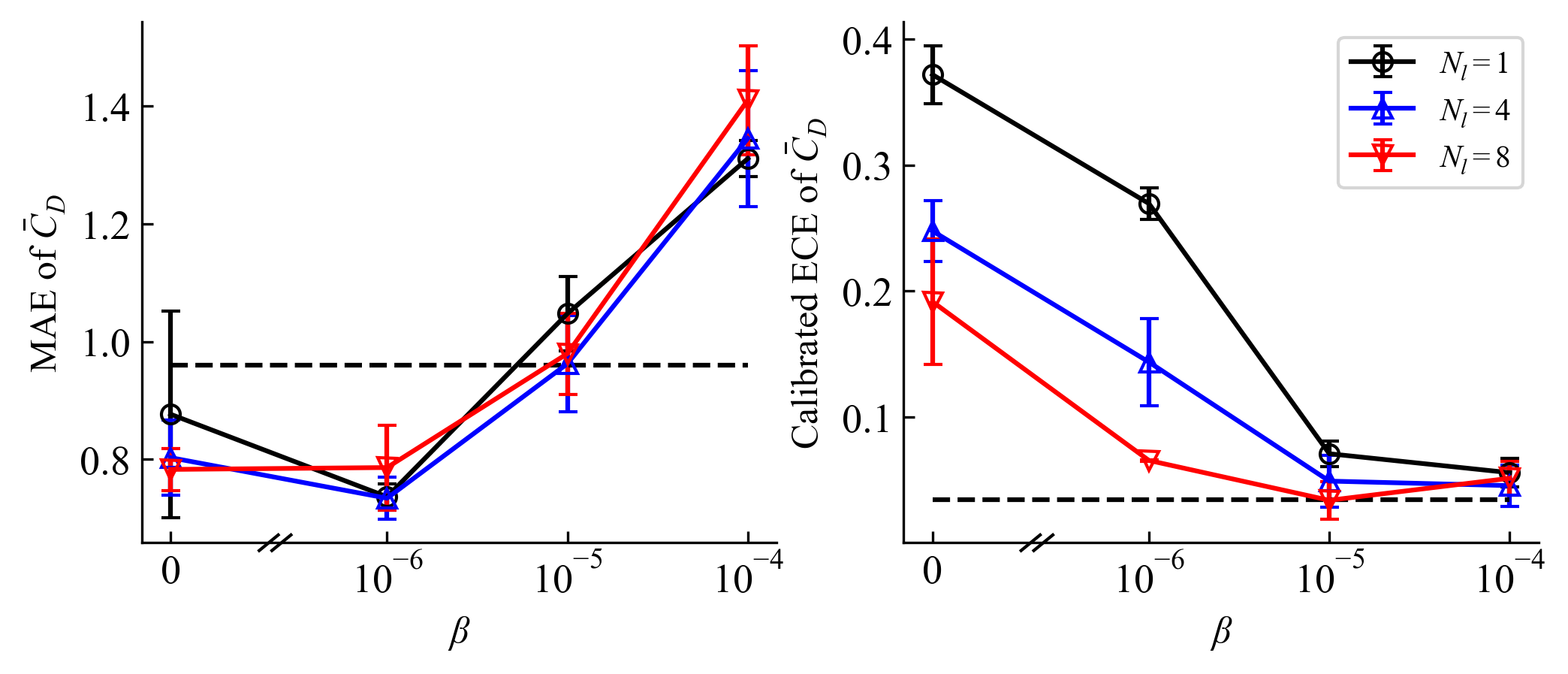}
    \caption{Model performance with different hyperparameters}
    \label{fig:hyper}
    \end{figure}

Following the evaluation procedure in Section \ref{sec:perf}, the model performance is assessed using the MAE and the calibrated ECE of $\bar C_D$, which quantify the accuracy of mean predictions and uncertainty estimates, respectively. For each hyperparameter configuration, three cross-validation runs are performed, and Fig. \ref{fig:hyper} reports the averaged results with error bars.

In Fig. \ref{fig:hyper}, the dashed line in the left subfigure represents the prediction error of the deterministic ED model, serving as a baseline reference for MAE. In the right, the dashed line corresponds to the calibrated ECE obtained from a deep ensemble method with $n=6$, providing a benchmark for high-quality uncertainty estimation.

The results in Fig. \ref{fig:hyper} reveal several important trends regarding the behavior of the GS-ED model under different uncertainty-related hyperparameter settings. For $\beta = 0$, where no additional KLregularization is applied, the model achieves low MAE across all values of $N_l$, indicating strong predictive performance. However, the corresponding calibrated ECE is significantly worse and remains far above the ensemble reference line, demonstrating that the resulting uncertainty estimates are poorly calibrated. As $\beta$ increases, the MAE gradually worsens, but the ECE consistently decreases, approaching the performance of the ensemble baseline at $\beta = 10^{-5}$ and $10^{-4}$. This opposite trend suggests that the KLterm acts as an effective distributional regularizer: although it slightly harms point-prediction accuracy, it substantially improves the reliability of the uncertainty estimates.

As for the sampling size $N_l$, the baseline case of $N_l = 1$ yields both higher MAE and higher calibrated ECE across most hyperparameters, particularly for small $\beta$. Increasing the sampling size to $N_l = 4$ produces consistent improvements, and further increasing to $N_l = 8$ provides only marginal additional benefits. This pattern indicates that enlarging the sampling size during training helps stabilize the learned predictive distribution.

Considering both prediction accuracy and the quality of uncertainty calibration, the configuration of $\beta = 10^{-5}$ with $N_l = 4$ provides a favorable balance, especially in downstream optimization tasks where uncertainty plays a critical role. This setting achieves an MAE close to that of the deterministic baseline while producing well-behaved uncertainty estimates that approach the ensemble benchmark, and serves as the model used in the main text and optimizations.

\section{Setup and results for buffet onset optimization}\label{app:buffet}
\setcounter{figure}{0}

In our previous study \citep{yang_fast_2024}, we conducted DBO for the buffet onset optimization, which is effective most of the time, but still results in failures in some cases. In this appendix, we extend UA-DBO to the same optimization problem to evaluate its performance.  

\subsection{Problem definition}

The two objectives are to maximize the airfoil's cruise lift-drag ratio, $(L/D)_\mathrm{cruise}$, and the buffet onset lift coefficient, $C_{L,\mathrm{buffet}}$. The buffet onset is predicted using the lift-curve-break criterion \citep{kenway_buffet-onset_2017, petrocchi_buffet_2022}, based on RANS-simulated lift coefficients at multiple $AOA$s. Specifically, the lift curve is generated from lift coefficients over a range of angles of attack, and buffet onset is identified by shifting the linear segment of the lift curve 0.1 degrees to the right and locating its intersection with the original curve. Constraints are applied to limit deviations in maximum thickness $t_{\max}$, cruise $AOA$, and pitching moment relative to the baseline. The 20 CST coefficients are still used to describe the airfoil shape, with an additional parameter for maximum airfoil relative thickness $(t/c)_{\max}$ to enhance control over thickness. The optimization problem can be expressed as follows:

\begin{align}
    \max_{\{u_i, l_i\}_{i = 0,\cdots, 9}, (t/c)_{\max}} &(L/D)_\mathrm{cruise},C_{L,\mathrm{buffet}} \\
    s.t. \quad & C_{L,\mathrm{cruise}} = C_{L,\mathrm{cruise}}^{(0)},\nonumber\\
    & 0.98 \cdot t_{\max}^{(0)} \le t_{\max}^{(0)} \le 1.02 \cdot t_{\max}^{(0)}, \nonumber \\
    & 0.95 \cdot t_{0.15c}^{(0)} \le t_{0.15c}\nonumber \\
    & AOA_{\mathrm{cruise}}^{(0)} - 0.5^{\circ} \le AOA_{\mathrm{cruise}} \le  AOA_{\mathrm{cruise}}^{(0)} + 0.5^{\circ}\nonumber\\
    & 0.9 \cdot C_{M,\mathrm{cruise}}^{(0)} \le C_{M,\mathrm{cruise}} \le 1.1 \cdot C_{M,\mathrm{cruise}}^{(0)}\nonumber
\end{align}




\subsection{Model setup}\label{app:model}

\begin{figure}[htbp]
    \centering
    \includegraphics[width=0.8\linewidth]{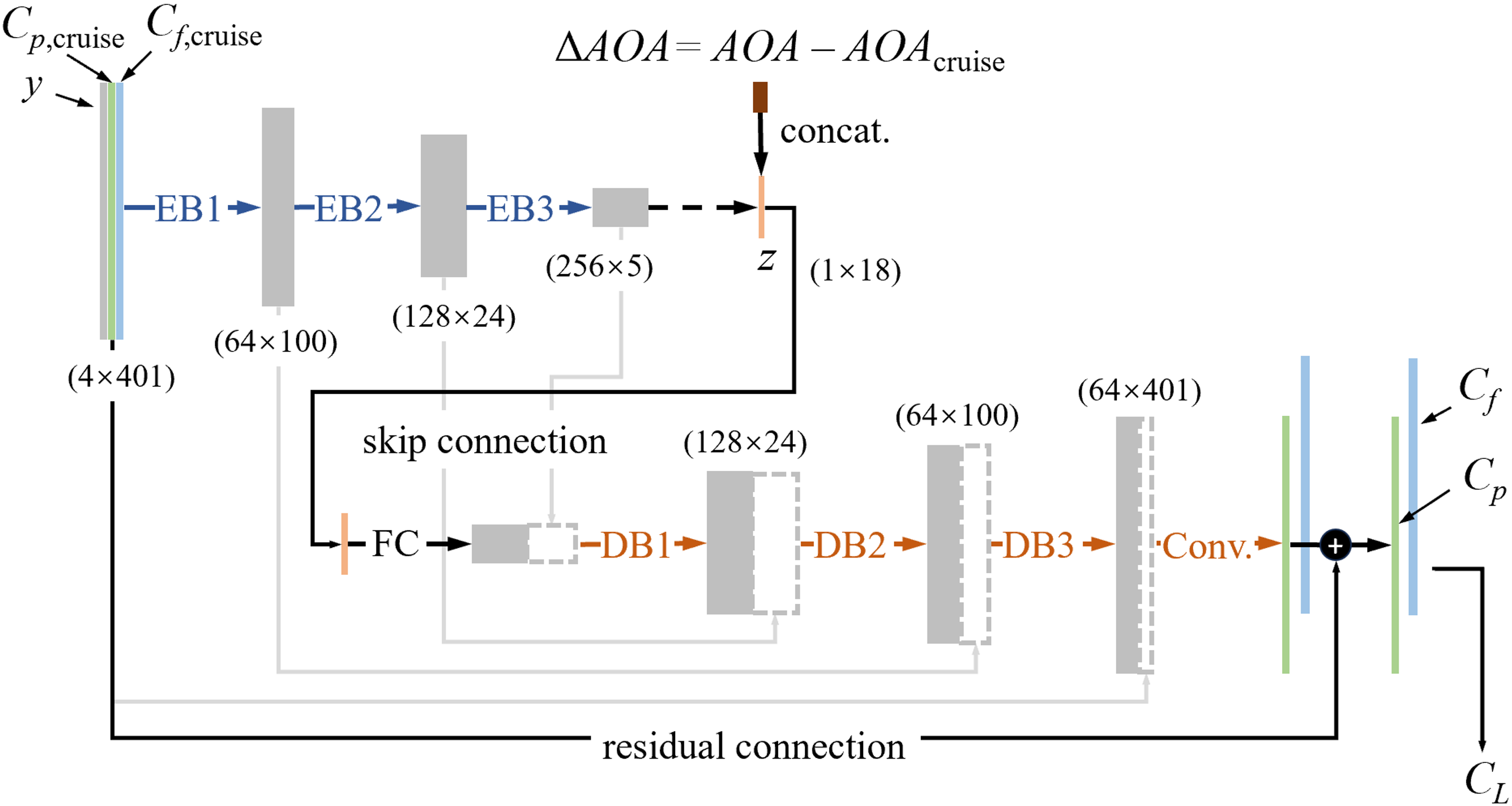}
    \caption{Architecture of the model for buffet onset optimization (The EB\# and DB\# stand for the encoder and decoder blocks, respectively. The FC and conv stand for the fully connected layer and the 1D Convolutional layer, respectively. The values in brackets indicate the channel number and feature map sizes of the tensors.)}
    \label{fig:p2mod}
\end{figure}

The model architecture is depicted in Fig. \ref{fig:p2mod}. It replicates the pUNet model from our previous study \cite{yang_fast_2024} to ensure a fair comparison between deterministic and probabilistic frameworks. The encoder and latent space remain identical to those in drag divergence optimization, and the target $AOA$ is concatenated to the latent vector. 

The model’s primary outputs are the surface $C_p$ and $C_f$ distributions under target conditions, which provide additional information critical for identifying buffet onset. The reason of using flow field rather than a scalar performance metrics as output was demonstrated in our previous work \cite{yang_fast_2024}. It improves generalization, interpretability, and task flexibility, which is particularly beneficial for complex phenomena such as buffet onset.

Specifically, two separate decoders predict the $C_p$ and $C_f$ distributions, each mirroring the encoder's structure but replacing the 1D convolutional layer with a linear interpolation layer for upsampling, followed by a convolutional layer with a stride of 1 for feature manipulation. Each decoder concludes with a final convolutional layer with a stride of 1 that compresses the output to a single channel.

The similarity between input and output shapes enables the use of a U-Net architecture, where skip connections link corresponding encoder and decoder feature maps to enhance the gradient flow. Additionally, residual learning is adopted: the decoder predicts the difference between the target and prior flow fields, rather than the target fields directly. Meanwhile, the input operating condition is also modified to the difference between the target and cruise $AOA$s.

\begin{figure}[H]
    \centering
    \includegraphics[width=0.35\linewidth]{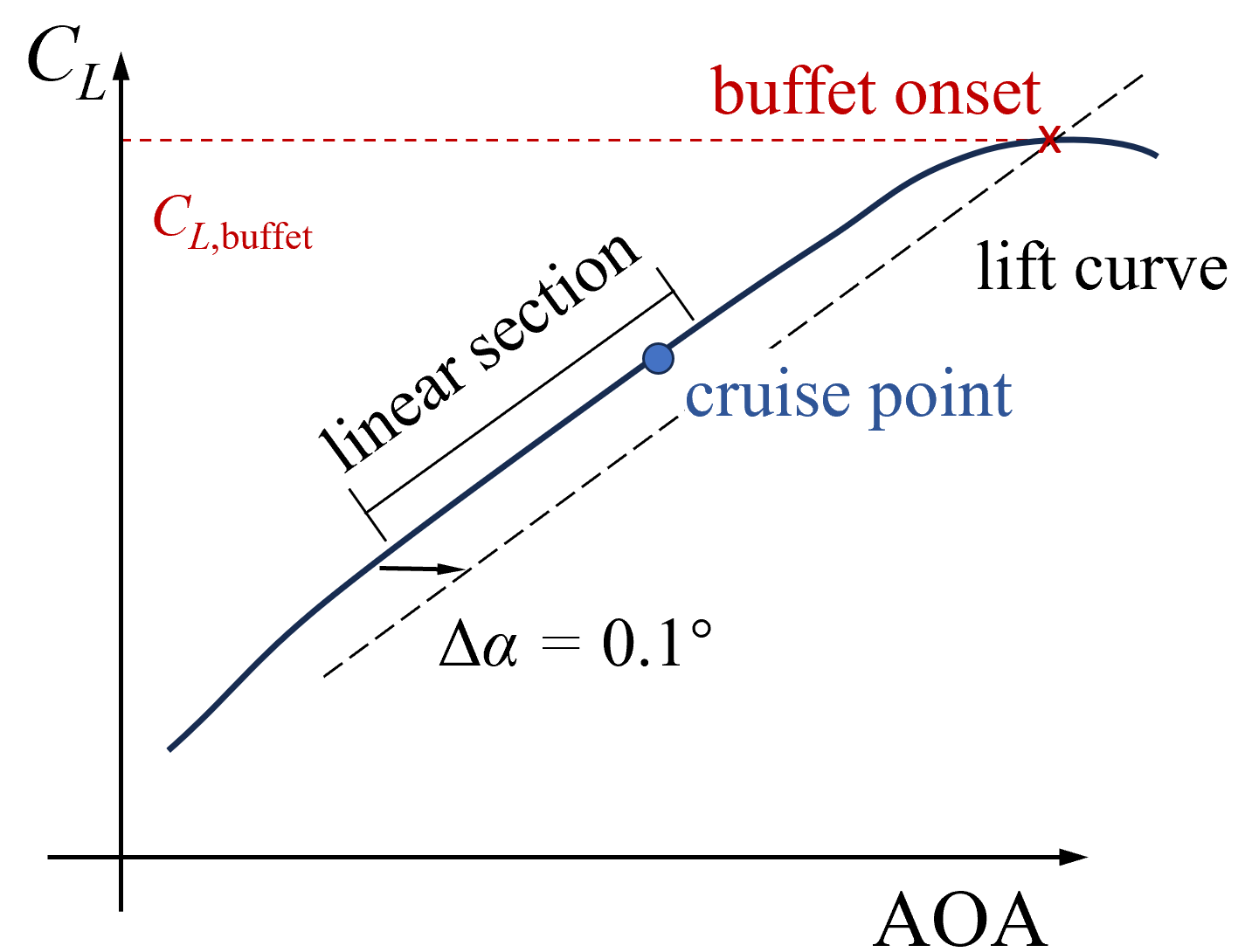}
    \caption{Lift-curve-break buffet onset criterion for buffet onset}
    \label{fig:buffet}
\end{figure}

For buffet onset prediction, the surface distributions are predicted with the model across multiple off-design angles of attack. Lift coefficients and other flow features are extracted from these distributions, and buffet onset is determined using the lift-curve-break criterion \citep{kenway_buffet-onset_2017}, as illustrated in Fig. \ref{fig:buffet}. Specifically, the lift curve is generated from lift coefficients over a range of angles of attack, and buffet onset is identified by shifting the linear segment of the lift curve 0.1 degrees to the right and locating its intersection with the original curve.

The model is trained with the same dataset as in our previous study  \cite{yang_fast_2024}. It has 25607 samples where off-design AOAs are sampled between $–3^\circ$ and $5^\circ$, with a constant Mach number for each airfoil. 

\subsection{Optimization setup}

\begin{figure}[htbp]
    \centering
    \includegraphics[width=0.9\linewidth]{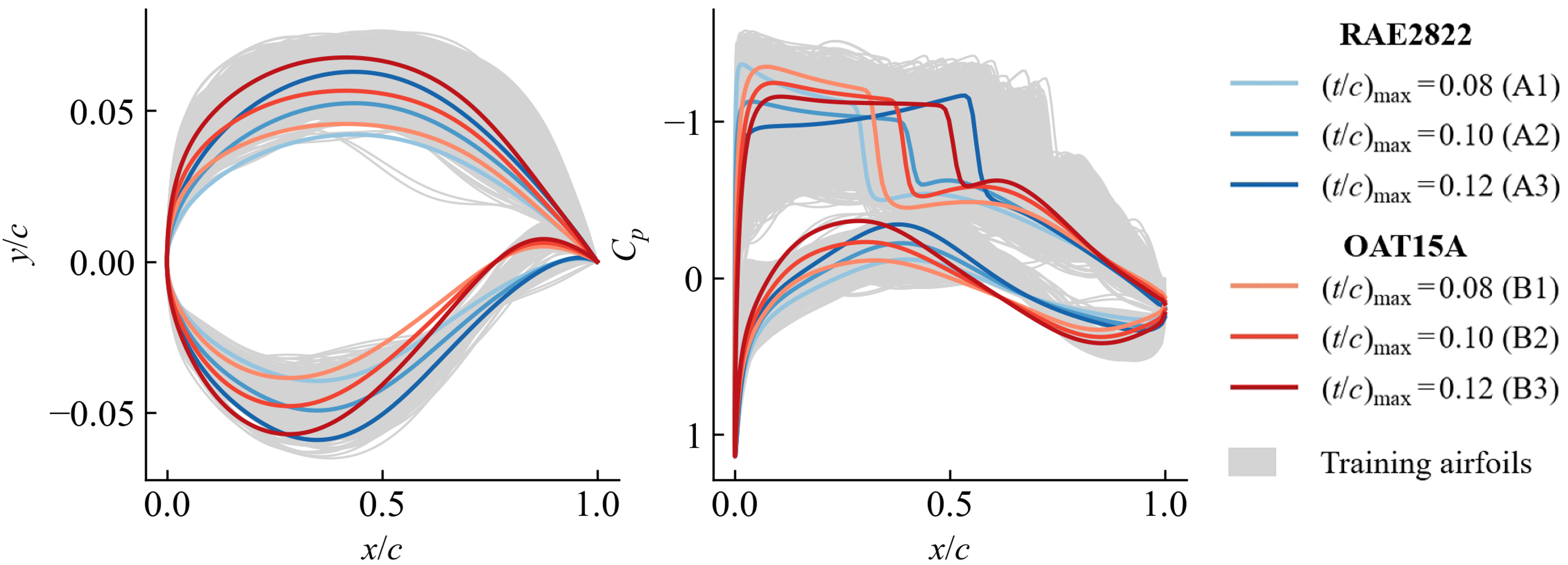}
    \caption{Geometries and cruise surface pressure distributions of baseline}
    \label{fig:airfoilp2}
\end{figure}

\paragraph{Baseline airfoils} In addition to baseline airfoils derived from RAE2822, another group is generated from the OAT15A airfoil: originally with a maximum relative thickness of 0.123, and is also rescaled to 0.083 and 0.103. Their geometries and cruise pressure coefficient distributions are shown in Fig. \ref{fig:airfoilp2}. The cruise conditions are $Ma_\infty = 0.734$ and $C_L = 0.70$ for the RAE2822 group, and $Ma_\infty = 0.730$ and $C_L = 0.75$ for the OAT15A group. 

\paragraph{Optimization algorithm} The two optimization objectives are the cruise lift-drag ratio and the buffet onset lift coefficient under cruise Mach conditions. AeroOPT is used as the optimization platform, initialized with 64 samples generated via bump perturbations, followed by 50 iterations with a population size of 32 per iteration.

\subsection{Optimization results}
 
\subsubsection{Performance of the probabilistic model in optimization}

\begin{figure}[htbp]
    \centering
    \begin{subfigure}{1\textwidth}
        \centering
        \includegraphics[width=0.8\linewidth]{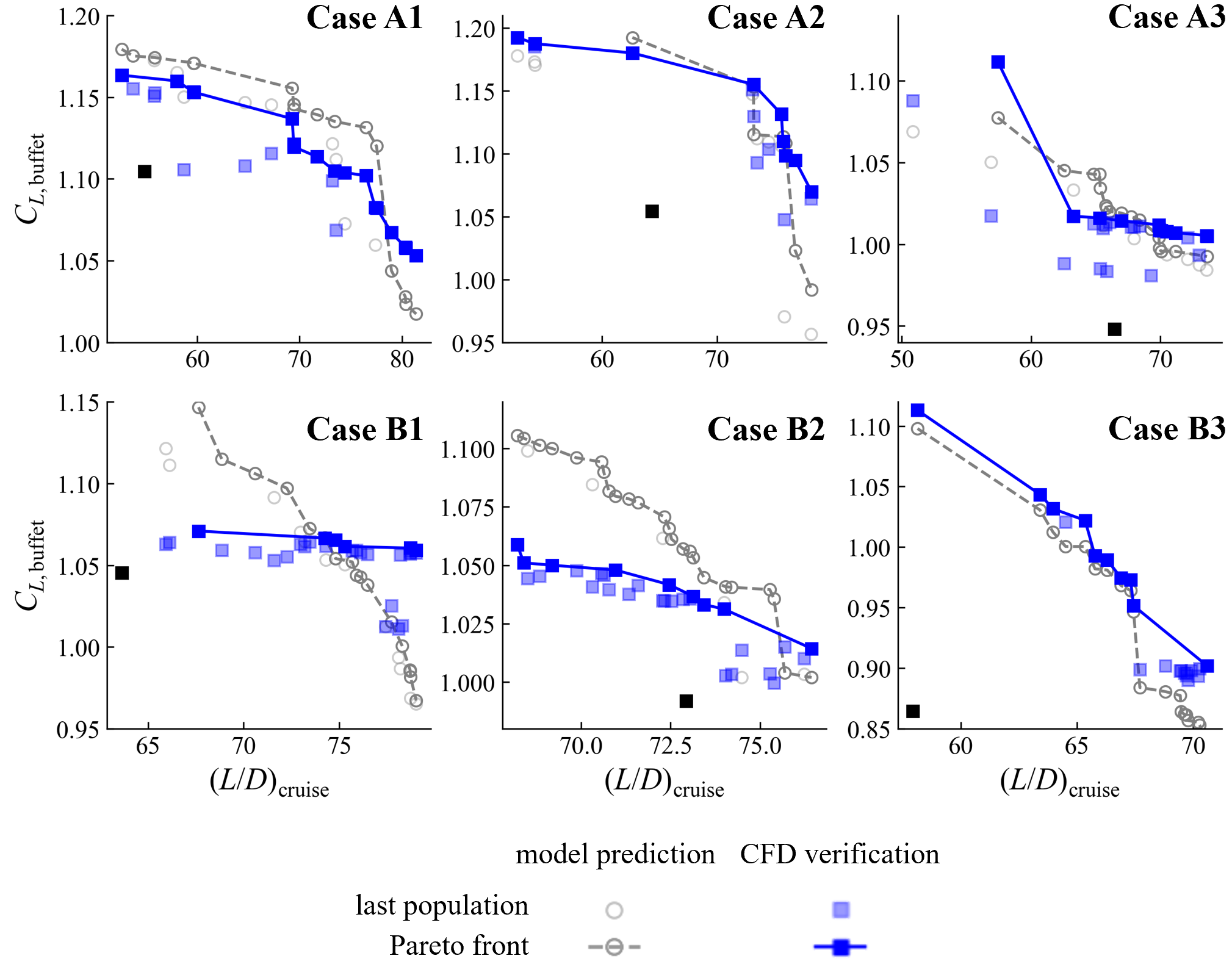}
        \caption{DBO}
        \label{fig:p2dbo}
    \end{subfigure}
    \hfill
    \begin{subfigure}{1\textwidth}
        \centering
        \includegraphics[width=0.8\linewidth]{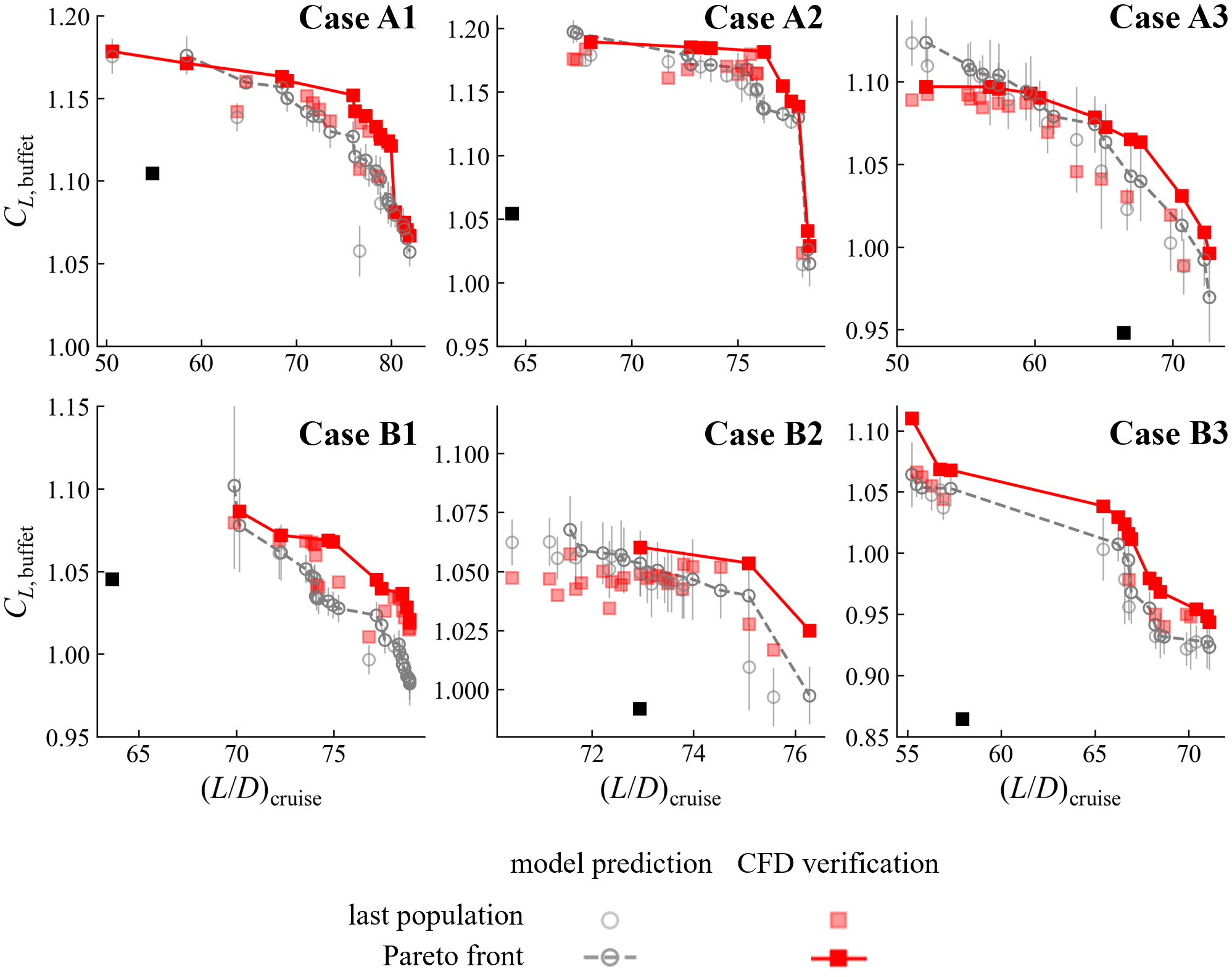}
        \caption{Mach number contours}
        \label{fig:p2uadbo}
    \end{subfigure}
    \caption{Comparison of Pareto fronts from original DBO and UA-DBO optimization}
    \label{fig:p2}
\end{figure}

Since a dual-objective optimization is conducted, the optimal solutions are expected to form a Pareto front balancing both objectives. To evaluate whether UA-DBO reduces model prediction errors, the errors associated with the Pareto-optimal samples across six cases are analyzed. Fig. \ref{fig:p2} provides an intuitive comparison: DBO results from \cite{yang_fast_2024} are shown in Fig. \ref{fig:p2dbo} and UA-DBO results in Fig. \ref{fig:p2uadbo}. Black squares denote the baseline performance. Gray circles represent the model-predicted performance in the final population, whose lift-drag ratios are accurate, but buffet onsets are predicted by models. The gray dashed lines represent the non-dominated Pareto front of the model-predicted buffet onset. Then, CFD verifications are conducted on each last population samples to obtain their actual buffet onset, which is marked by solid squares, with the actual Pareto fronts shown as solid lines.

In Fig. \ref{fig:p2dbo}, significant prediction errors are observed in several cases (e.g., Cases B1 and B2), where the dashed actual Pareto fronts are below the model-predicted ones, indicating over-predictions of buffet onset. In contrast, such discrepancies are mitigated in UA-DBO results. The red dashed and solid lines in Fig. \ref{fig:p2uadbo} closely match across all optimization cases, indicating improved predictive accuracy.

Moreover, UA-DBO consistently yields samples with superior CFD-verified buffet onset lift coefficients. An important observation from Fig. \ref{fig:p2} is that while the model-predicted Pareto fronts for DBO and UA-DBO (gray dashed lines) are similar, the actual Pareto fronts for UA-DBO consistently achieve higher buffet onset values. This demonstrates that enhancing model reliability can improve optimization outcomes.

A quantitative comparison of model prediction errors, measured by the MAE of buffet onset lift coefficients in the final populations, is provided in Table \ref{tab:opt2}. The values in parentheses beside the UA-DBO columns denote the percentage reduction relative to the original DBO framework, calculated as the reduction divided by the original value. On average, UA-DBO reduces model prediction errors by 32.8\%.

\begin{table}[ht]
\centering
\caption{Model prediction errors on the last population for DBO and UA-DBO}\label{tab:opt2}%
\begin{tabular*}{0.6\textwidth}{@{\extracolsep\fill}c|cc}
\toprule
 Case &   DBO \citep{yang_fast_2024} & UA-DBO\\
\midrule
 A1\footnotemark[1] & 0.027 & 0.016 (–38.4\%) \\
  A2 & 0.017 & 0.016 (–7.6\%) \\
  A3 & 0.019 & 0.013 (–30.5\%) \\
 \midrule
 B1\footnotemark[1] & 0.037 & 0.024 (–34.9\%) \\
  B2 & 0.032 & 0.010 (–70.3\%) \\
  B3 & 0.028 & 0.024 (–15.1\%) \\
\bottomrule
\end{tabular*}
\footnotetext[1]{The maximum relative thickness of the baseline airfoil is beyond the training dataset}
\end{table}

\subsubsection{Performance of optimized airfoils}

Additionally, for each optimization case, a representative sample from the DBO and UA-DBO Pareto fronts is selected, corresponding to the largest buffet onset improvement while maintaining a cruise lift-drag ratio exceeding the baseline. Their CFD-verified buffet onset lift coefficients and cruise lift-drag ratios are presented in Table \ref{tab:opt21}. The percentages in parentheses represent improvements relative to the baseline, and the results that UA-DBO outperformed original DBO are marked with bold. In four out of six cases, the representative UA-DBO samples demonstrate superior performance in both objectives compared to their DBO counterparts, and in the remaining two cases, at least one objective shows improvement. These results affirm the capability of UA-DBO to consistently yield better results.

\begin{table}[ht]
\centering
\caption{CFD-simulated performance of the typical sample on the Pareto front (the results that UA-DBO outperformed DBO are marked bold)}\label{tab:opt21}%
\begin{tabular*}{\textwidth}{@{\extracolsep\fill}cccccccc}
\toprule
& case & \multicolumn{2}{c}{baseline} & \multicolumn{2}{c}{typical sample of DBO} & \multicolumn{2}{c}{typical sample of UA-DBO} \\
\cmidrule{3-4}\cmidrule{5-6}\cmidrule{7-8}
 & & $C_{L,\mathrm{buffet}}$  & $(L/D)_\mathrm{cruise}$ & $C_{L,\mathrm{buffet}}$  & $(L/D)_\mathrm{cruise}$ & $C_{L,\mathrm{buffet}}$  & $(L/D)_\mathrm{cruise}$ \\
\midrule
\multirow{3}{*}{} & A1\footnotemark[1] & 1.10 & 54.8 & \makecell{1.16 \\ (5.0\%)} & \makecell{58.0 \\ (5.8\%)} & \makecell{\textbf{1.17} \\ \textbf{(6.0\%)}} & \makecell{\textbf{58.4} \\ \textbf{(6.6\%)}} \\
 & A2 & 1.05 & 64.4 & \makecell{1.16 \\ (9.6\%)} & \makecell{73.2 \\ (13.7\%)} & \makecell{\textbf{1.16} \\ \textbf{(10.5\%)}} & \makecell{\textbf{75.9} \\ \textbf{(17.9\%)}} \\
 & A3 & 0.95 & 66.5 & \makecell{1.01 \\ (7.0\%)} & \makecell{67.0 \\ (0.8\%)} & \makecell{\textbf{1.02} \\ \textbf{(7.5\%)}} & \makecell{\textbf{69.8} \\ \textbf{(5.1\%)}} \\
\multirow{3}{*}{} & B1\footnotemark[1] & 1.05 & 63.6 & \makecell{1.07 \\ (2.4\%)} & \makecell{67.6 \\ (6.3\%)} & \makecell{\textbf{1.09} \\ \textbf{(3.9\%)}} & \makecell{\textbf{70.1} \\ \textbf{(10.2\%)}} \\
 & B2 & 0.99 & 72.9 & \makecell{1.04 \\ (4.5\%)} & \makecell{73.1 \\ (0.3\%)} & \makecell{\textbf{1.06} \\ \textbf{(6.9\%)}} & \makecell{73.0 \\ (0.0\%)} \\
 & B3 & 0.86 & 57.9 & \makecell{1.11 \\ (28.8\%)} & \makecell{58.1 \\ (0.4\%)} & \makecell{1.04 \\ (20.1\%)} & \makecell{\textbf{65.4} \\ \textbf{(12.9\%)}} \\
\bottomrule
\end{tabular*}
\footnotetext[1]{The maximum relative thickness of the baseline airfoil is beyond the training dataset}
\end{table}

\begin{figure}[htbp]
    \centering
    \includegraphics[width=0.9\linewidth]{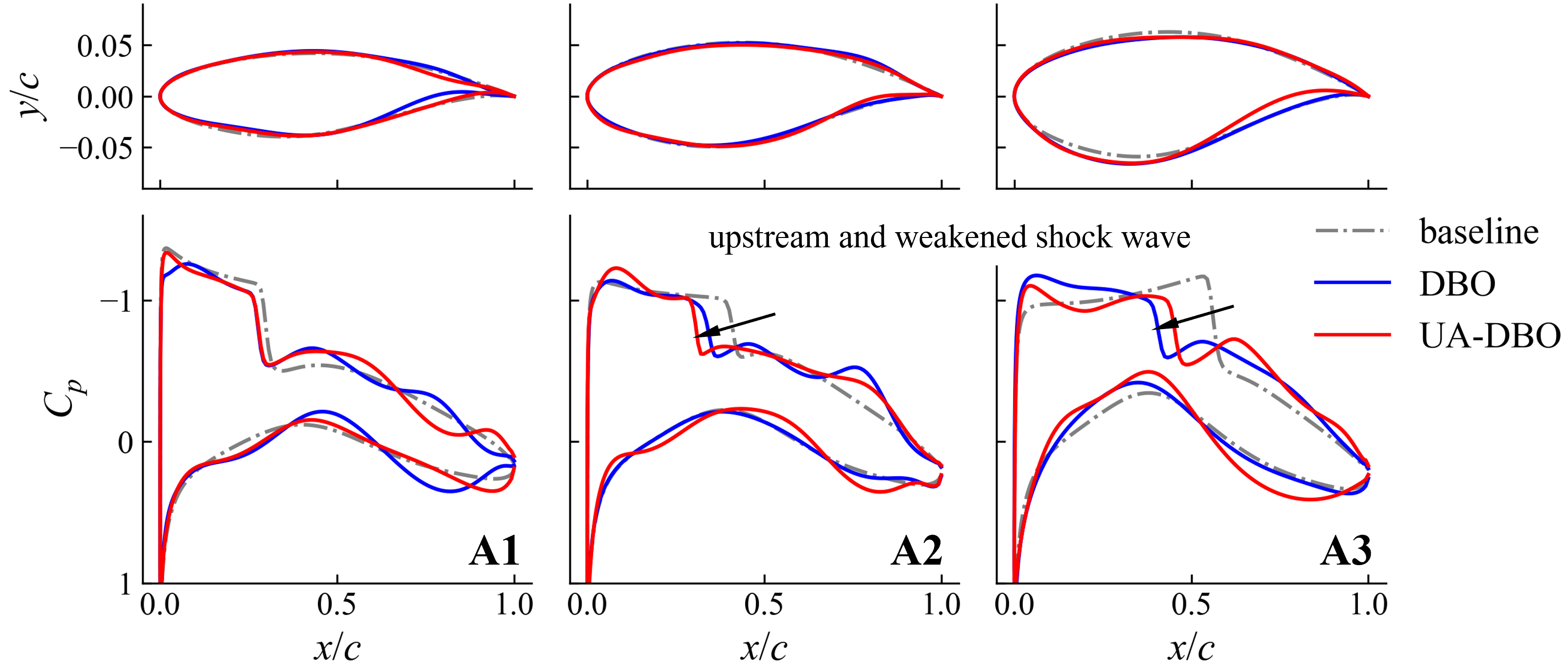}
    \includegraphics[width=0.9\linewidth]{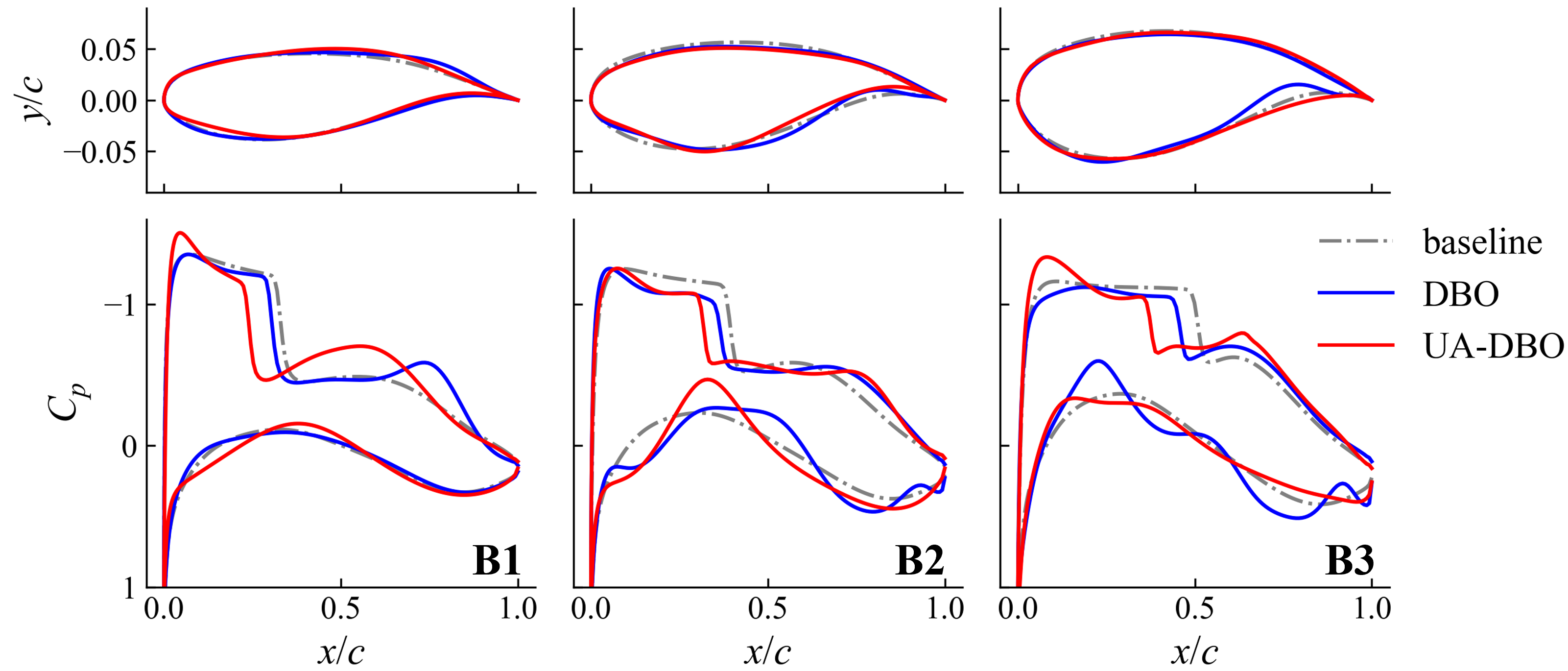}
    \caption{Typical optimized airfoil geometries and cruise pressure coefficient distributions}
    \label{fig:p2res}
\end{figure}

The typical optimized airfoil geometries and their CFD-simulated Cp distributions under cruise conditions are shown in Fig. \ref{fig:p2res}. The optimized airfoils from UA-DBO and DBO exhibit trends, with shock waves on the upper surfaces shifted further upstream and weakened in intensity, both favorable for delaying buffet onset. Additionally, UA-DBO promotes the generation of samples with smoother $C_p$ distributions, reducing unfavorable bumps and fluctuations during cruise conditions.

\end{appendices}


\bibliographystyle{unsrt}
\bibliography{sn-bibliography}

\end{document}